# How to improve the interpretability of kernel learning

Jinwei Zhao[a,b,+], Qizhou Wang[a,b], Yufei Wang[a,b], Yu Liu[a,b], Zhenghao Shi[a,b], Xinhong Hei[a,b,+]

[a]School of Computer Science and Engineering, Xi'an University of Technology, Xi'an 710048, China

[b]Shaanxi Key Laboratory of Network Computing and Security Technology (Xi'an University of Technology), Xi'an 710048, China

+ Corresponding author: Jinwei Zhao, Xinhong Hei

E-mail: zhaojinwei@xaut.edu.cn, xinghonghei@xaut.edu.cn

Phone: +86-029-82312196(2016)

**Abstract**: In recent years, machine learning researchers have focused on methods to construct flexible and interpretable prediction models. However, an interpretability evaluation, a relationship between generalization performance and an interpretability of the model and a method for improving the interpretability have to be considered. In this paper, a quantitative index of the interpretability is proposed and its rationality is proved, and equilibrium problem between the interpretability and the generalization performance is analyzed. Probability upper bound of the sum of the two performances is analyzed. For traditional supervised kernel machine learning problem, a universal learning framework is put forward to solve the equilibrium problem between the two performances. The condition for global optimal solution based on the framework is deduced. The learning framework is applied to the least-squares support vector machine and is evaluated by some experiments.

## 1. Introduction

Safe, controllable and credible artificial intelligence has been the goal which the humanity has been pursuing. In the field of machine learning, in order to achieve this goal, it is necessary for learning algorithm to really interact with the humanity; It is necessary for the learning algorithm to have the ability to correct errors, so as to avoid a prediction model with serious errors caused by unnecessary deviation in training data; It needs to be able to check its own learning process or decision-making process based on unsuccessful prediction results, especially for complex learning tasks; It is necessary to establish a learning algorithm for capturing and learning causal relationships in the world around us, so that the prediction model could predict what will happen under certain conditions, even if these conditions are significantly different from those of the past; It needs the learning algorithm which can really take full control of generalization performance of the prediction model. As big data accelerates transformation of scientific research pattern, scientific research is translating from a hypothetical drive mode to a data-driven one, which needs learning algorithm to discover new natural phenomena and laws through big data mining, statistic and analysis. However, recently, all of this is out of reach. The reason is that the prediction model and its training process are not yet understood by human beings, and are not covered by the knowledge base we currently own.

In practice, occurrence of random events always causes deviation of measurement data, which generates many intermittent or continuous noise data, so as to make the prediction model deviated from known relationship and real law between data. Meanwhile, because the training data set is just a subset in sample space, if we can't discover its distribution, as well as sample size is not large enough, even if there is no noise data, finally the prediction model can't accurately express real relationship and law between data. Even if enough data were collected, if they do not conform to its real distribution, the final prediction model will be the same as the result from usual small sample. Even if distribution is known, but if the basis of the prediction model's space is not known in advance, traditional machine learning algorithms still can't guarantee the final prediction model can express exactly the true relationship and the real law between data and difficultly ensure the model full compliance with professional knowledge. We believe this is because that the linear functional set from conjugate space of high dimensional feature space obtained by kernel method is nowhere dense in the square integrable function space. Even if mathematical form of the prediction model is known in advance, if the optimization problem is multi-peak complex objective function, recently there is no strong optimization mechanism to solve this optimization problem effectively for the optimal interpretable prediction model. In order to achieve this goal, the sample must be dense enough, its distribution must be accurately known, and interference of the noise data can be avoided easily, the kernel function should be reasonable or mathematical expression of the prediction model must be fully known, even prediction model posterior distribution is clear and a good optimization algorithm is also essential. However, in fact, all learning algorithms we face do not have such strict prerequisites.

How to make the prediction model and training process understood by us, and conform to human cognition, even to generate new cognitive for human, is in essence an optimization problem which can promote the interpretability of the prediction model and make the model more suitable to its causality or discover faults in the causality. As literature [1] points out, when we pay attention to scientific problems, it is a motivation of scientific research to trace their origins, or to pursue their causality. Professor Zoubin Ghahramani also pointed out that current machine learning theorists should consider how to construct more flexible and interpretable prediction models[2]. In ICML 2017, the theme of the best paper "Understanding black-box Predictions via Influence Functions" is to use influence functions to understand black-box predictions and study how to explain source of prediction models[3]. Many researchers have presented their results on NIPS'17 Interpretable Machine Learning Symposium and on CVPR Tutorial 2018 Interpretable Machine Learning for Computer Vision. FICO, Google, UC Berkeley, Oxford, Imperial, MIT and UC Irvine jointly launched a competition to generate new researches for the interpretability of machine learning algorithms.

The interpretability of the prediction model is generally regarded as human imitativeness. If humans can explain every calculation steps and finally make prediction at right time by using input data and model parameter, the prediction model will have this kind of imitativeness which is the interpretability (Lipton, 2016). For example, given a simulation model for a diagnosis, a doctor can easily check each step of the model with their professional knowledge and even infer fairness and system deviation of the diagnosis result. However, this is a strict definition. If based on this definition to improve the interpretability, the domain knowledge must be forced to every step of the training process of the prediction model. The optimal prediction model tends to lose its generalization performance, such as decision tree algorithm.

We posit that the interpretability should be a potential ability to help experts discover an essential reason of a prediction result and provide research clues and possibilities for researchers to further research. Specifically, that is, when the prediction model is the same as a theoretical model in the form of geometric shape or mathematical expression, despite different value and different scale, we can think that the model has good interpretability, that the model can be explained well by the theoretical model.

We posit that in machine learning, it is more realistic to apply this definition to solve an interpretability improvement problem of the prediction model. Its key problem is how to ensure the prediction model as much as possible consistent with its explanatory description, but not lost its generalization performance. In a word, in a training process of the prediction model, we not only should consider its generalization performance but also need to consider the deviation between the prediction model and the theoretical model, namely the interpretability.

Currently, there are two methods for improving the interpretability of the prediction model: analytical interpretability and statistical interpretability. In the analytical interpretability, Pang et al. [3], Wu et al. [4], and Zhou et al. [5-6] respectively discovered and analyzed the prediction model by influential function and visualization of internal feature data of neural networks. Craven et al.[7], and Baehrens et al.[8] proposed model-agnostic method by learning an interpretable model on the predictions of the black box model. Strumbelj et al.[9] and Krause et al. [10] proposed perturbing inputs and seeing how the black box model reacts.   In statistical interpretability, James et al.[11] proposed to build a prediction model of automatic statistical learning by mining the functional relationship between input and output from training samples. Ribeiro et al.[12-13] proposed a measure of the interpretability complexity for obtaining an interpretable linear model and realizing a local interpretability and proposed a submodular pick algorithm for a global interpretability. We in a literature [14] proposed use of prior knowledge in a hypothesis space (such as Sobolev space) to construct a compact subset, which can ensure the interpretability of the prediction model and correct prior knowledge in its training process.

However, how is the interpretability of the prediction model evaluated? How does promotion of its interpretability affect its generalization performance? If there is an equilibrium problem between the two, what is the probability upper bound for the sum of the two? What factors are involved in this upper bound? What are the relationship between the factors and the sum? What is the learning framework for solving this equilibrium problem? Is there a unique solution to this optimization problem? What are the conditions for producing the unique solution? These questions have not yet been answered. For this purpose, this paper first introduces the learning framework of traditional machine learning based on $L_2$ norm regular term. Then, the quantitative index of the interpretability is proposed and its rationality is given, and the relationship between the interpretability and the generalization performance is analyzed. Probability upper bound of the sum of the two performances is analyzed. For traditional supervised kernel machine learning problem, a universal learning framework is put forward to solve the equilibrium problem between the two performances. The uniqueness of solution of the problem is proved and condition of unique solution is obtained.

## 2. Learning framework of traditional machine learning

Suppose $X$ is a compact domain or a manifold in Euclidean space and $Y \in R^k$, $k = 1$, $\rho$ is a Borel probability measure of a space $Z = X \times Y$.

$f_\rho: X \to Y$ as $f_\rho(x) = \int_Y y \mathrm{d}\rho(y|x)$ is defined. The function $f_\rho$ is a regression function of $\rho$.

In machine learning, $\rho$ and $f_\rho$ are unknown. At some conditions, an edge probability measure $\rho_X$ of $X$ is known.

The goal of the learning is to find the best approximation of $f_\rho$ in a square integrable function space $\mathcal{H}_K$ spanned by a kernel function $K$. The number of sample data is $m$. Therefore, Tihonov regularization learning framework[23-24] can be obtained

$$f_{\mathbf{z},\gamma} = \mathrm{argmax}_{f \in \mathcal{H}_K} \left\{ \frac{1}{m} \sum_{i=1}^m (f(x_i) - y_i)^2 + \gamma \|f\|_K^2 \right\} \quad (1)$$

# 3. Interpretability of the prediction model

Kernel machine learning method [15] improves the generalization performance of the prediction model based on traditional machine learning framework by analyzing generalization error bound [16]. However, if trying to ensure that the prediction model can be explained as far as possible and do not lose its generalization performance in the learning process, not only the generalization error bound should be considered, but also the deviation boundary between the prediction model and a mathematical model describing prior knowledge, denoted by interpretability model, need to be considered too. Literature [2] proposed a uniform description of all optimization problems based on prior knowledge by assuming that the interpretability model and the prediction model are in the same Hilbert space and finished consistency analysis and error analysis by using the prior knowledge as strong constraints. However, the following two problems have not been well solved: 1. How to quantify the interpretability of the prediction model? 2. When the interpretability model and the prediction model are not in the same Hilbert space, how to use prior knowledge to constrain the learning process for a well interpretability. This section will be the first to put forward a quantitative evaluation index of the interpretability, and then prove the existence of an equilibrium problem between the generalization performance and the interpretability based on the index.

## 3.1 Evaluation of the interpretability of the prediction model

The interpretability of the prediction model is not innate, and should be given by domain experts based on professional terms or common sense, such as prior knowledge. Meanwhile, this kind of professional explanation should be expressed in terms of mathematical functions, denoted by interpretation function or interpretation model, such as linear models[9, 17], gradient vector[18], an additive model[19], decision trees[20], falling rule lists[21-22], attention-based networks and so on. However, the prior knowledge is usually uncertain and incomplete, which will lead to the uncertainty of the interpretation model. The method for expressing the uncertain and incomplete prior knowledge and for obtaining complete knowledge from incomplete knowledge, in another article, had been introduced. So this article starts with the assumption that the complete knowledge has been obtained.

Inspired by induction and analysis coupling learning method, the differences between the prediction model $f(x)$ and the interpretation function $P(x)$ can be computed by the mean square error between the two. However, in practice, the mean square error is too strict in evaluating the difference between the two models, and there will be an ill-posed problem caused by different magnitudes of the both functions and different function subspace. We posit that the interpretability of the prediction model itself depends on correct expression of causal relationship between the output attribute and the input attributes. When the attributes in both models satisfy the same causal relationship, we can assume that both models express the same interpretation, even though their magnitude is different. This conclusion can be explained by Fig. 1.

In Fig.1, there are three functions, $f(x) = ax$, $f_1(x) = bx$ and $f_2(x) = ax + c$. It is obvious when $c \gg \frac{1}{\sqrt{3}}|a-b|(x_2^2 + x_1 x_2 + x_1^2)^{\frac{1}{2}}$, the mean square error between $f_2(x)$ and $f(x)$ is $\int_{x_1}^{x_2}(f_2(x) - f(x))^2 dx = c^2(x_2 - x_1)$, which is greater than $\int_{x_1}^{x_2}(f(x) - f_1(x))^2 dx = \frac{1}{3}|a-b|^2(x_2^3 - x_1^3)$ which is the mean square error between $f_1(x)$ and $f(x)$. In order of magnitude, $f_1(x)$ is similar to $f(x)$, but the function relationship between the input and output attributes in $f_2(x)$ and $f(x)$ is consistent. So the mean square error can't compare difference of the function relationship between the input attributes and the output attribute. In the example, the variance of the error between $f_2(x)$ and $f(x)$ is exactly 0. Thus it can be seen that the variance of the error is better for evaluating the difference between the difference function relationships between the input attributes and the output attribute.

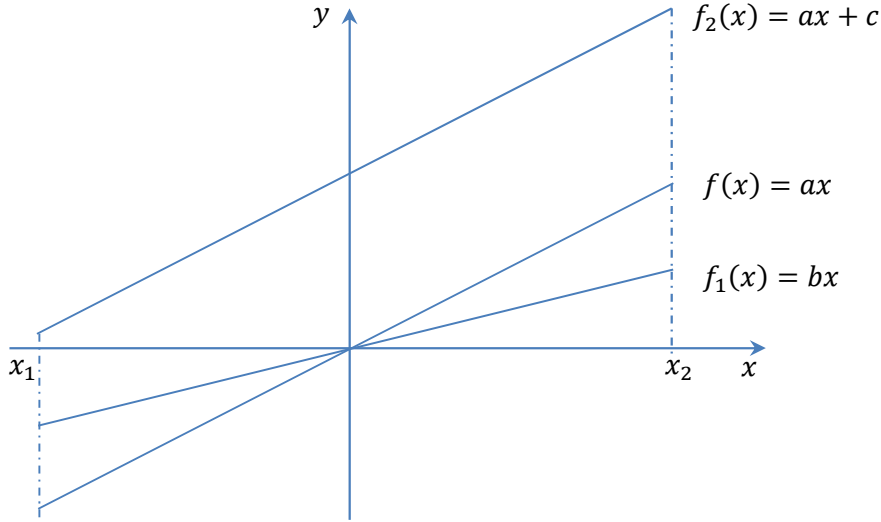

Fig.1 Two different evaluative methods of the interpretability

If $X$ is a compact metric space, $v$ is a Borel measure in $X$, such as Lebesgue measure or edge measures, $\mathcal{L}_v^2(X)$ is a square integrable function space on $X$. From the above discussion, it can be seen that in the function space $\mathcal{L}_v^2(X)$ the variance $\mathcal{E}^P(f)$ of the error between a model $f(x)$ and an interpretation model $P(x)$ can be used to calculate the interpretability of $f(x)$, also known as interpretation distance.

$$\mathcal{E}^P(f) = \int_Z \left(f(x) - P(x) - \mu^P(f)\right)^2 dv \qquad (2)$$

where

$$\mu^P(f) = \int_Z |f(x) - P(x)| \, dv \qquad (3)$$

is a mean error between $f(x)$ and P(x).

## 3.2 Sufficient and necessary conditions for consistent convergence of expected risk and interpretation distance

If the expected risk minimization can guarantee the minimum of the interpretation distance, traditional machine learning algorithm can improve the interpretable performance of the prediction model. If not, contradiction between the generalization performance and the interpretable performance will exit. A new machine learning frame structure should be needed to solve the equilibrium problem between the both performances. The essence of the problem is to find sufficient and necessary conditions for consistent convergence of expected risk and interpretation distance.

Firstly, the definition of consistent convergence should be given.

If $K: X \times X \to \mathbb{R}$ is a continuous function, an integral transform $(L_K f)(x) = \int K(x,t) f(t) dv(t)$ are a linear mapping: $L_K: \mathcal{L}_v^2(X) \to \mathcal{L}_v^2(X)$. The function $K$ is a kernel function of $L_K$. Let $K_x: X \to \mathbb{R}$ become $K_x(t) = K(x,t)$. The Hilbert-Schmidt theorem shows that if all eigenvalues of the operator $L_K$ are not strictly nonzero and $\phi_k(x)$ is an eigenfunction of $L_K$, $k = 1,2,3,\ldots$, in the Hilbert space $\mathcal{L}_v^2(X)$ any function $f(x)$ has $f(x) = \sum_k c_k \phi_k(x) + \xi(x)$, and $\xi(x) \in Ker L_K$ which is $L_K \xi(x) = 0$. $\xi(x)$ usually can be approximated by a bias. The kernel function $K$ can span a reproducing kernel Hilbert spaces, $\mathcal{H}_K = \left\{ f \in \mathcal{L}_v^2(X) \middle| f = \sum_{k=1}^{\infty} a_k \phi_k(x), with \left( \frac{a_k}{\sqrt{\lambda_k}} \right) \in \ell^2 \right\}$, which became a solution space of the kernel machine learning algorithm. In the space, the inner production is defined as $\langle f, g \rangle_K = \sum_{k=1}^{\infty} \frac{a_k b_k}{\lambda_k}$, where $f = \sum_{k=1}^{\infty} a_k \phi_k(x)$ and $g = \sum_{k=1}^{\infty} b_k \phi_k(x)$. $\mathcal{H}_K$ can be regarded as a linear function set on $\mathcal{L}_v^2(X)$. It is in this solution space that we will define the consistent convergence of expected risk and interpretation distance in $\mathcal{H}_K$.

**Definition 1** We say that the converge of expected risk and the converge of interpretation distance are consistent for the set of functions $\mathcal{H}_K$ and if the following two sequences converge to their minimum values in probability on the same function $f_z^*$ with the increase of the size of the training sample set:

$$\mathcal{E}^P(f_z) \xrightarrow[|z| \to \infty]{} \min_{f_z^* \in \mathcal{H}_K} \mathcal{E}^P(f_z^*) \tag{D1}$$

$$\mathcal{E}(f_z) \xrightarrow[|z| \to \infty]{} \min_{f_z^* \in \mathcal{H}_K} \mathcal{E}(f_z^*) \tag{D2}$$

The necessary and sufficient conditions for consistent convergence are discussed below.

If there is a sample set including $m$ samples, $\{(x_1, y_1), \cdots, (x_m, y_m)\}$, which is given by random sampling based on probability measure $\rho$ on $X \times Y$(This measure $\rho$ is usually unknown). On every $x \in X$, $\rho(y|x)$ is a condition probability measure of $y$ given $x$. $\rho_x$ is a marginal probability measure on $X$. $f_\rho(x)$ is the function that needs to be predicted on $X$ with the measure $\rho_x$, which is a mean value of $y$ on $\{x\} \times Y$ as $f_\rho(x) = \int_Y y d\rho(y|x)$.

Obviously, if there are the function $f_\rho(x)$ and $P(x)$ in $\mathcal{H}_K$, the converge is consistent if and only if $\mathcal{E}^P(f_\rho) = 0$. In by now, when the function $f_\rho(x)$ and $P(x)$ exist in $\mathcal{H}_K$ the sufficient and necessary conditions for the consistent convergence is $\mathcal{E}^P(f_\rho) = 0$.

The following discussion focuses on the absence of $f_\rho(x)$ or $P(x)$ in $\mathcal{H}_K$.

The function $f_\rho(x)$ usually approximate a linear functional in $\mathcal{H}_K$. And the interpretation function $P(x)$ can approximate a linear functional or exist in $\mathcal{H}_K$ (when all eigenfunctions of the operator $L_K$ are strictly nonzero). Likewise, the convergence of expected risk and interpretation distance is consistent if and only if $\mathcal{E}^P(f_\rho) = 0$. When $P(x)$ don't approximate any linear functional or don't exist in $\mathcal{H}_K$ and cannot be a linear weighted sum of all eigenfunctions of the operator $L_K$, it is a nonlinear functional on $\mathcal{L}_\nu^2(X)$. First, lemma 1 is given.

**Lemma 1.** A continuous linear functional set on a separable Hilbert space $X$ is nowhere dense in an integrable function space $\mathcal{L}_\nu^1(X)$.

**Proof:** The limitation of any nonlinear functional sequence is a simple function with finite values in the sense of average convergence, but it does not have features of linear function, such as, additive and homogeneous. If $M$ is a measurable set of the separable Hilbert space

$X$, and $v(M) < \infty$, based on the conditions of lebesgue measure (In real number set $R$ every open set and every closed set are measurable, and for all measurable set $M \subset R$ and arbitrary $\varepsilon > 0$ there is an open set $G \supset M$ and $v(G) - v(M) < \varepsilon$ can be obtained) we can obtain that for any $\varepsilon > 0$ a closed set $F_M$ and an open set $G_M$ can be found and have

$$F_M \subset M \subset G_M \text{ and } v(G_M) - v(F_M) < \varepsilon \tag{4}$$

A function $\varphi_\varepsilon(x)$ is defined as the following:

$$\varphi_\varepsilon(x) = \frac{\gamma(x, R - G_M)}{\gamma(x, R - G_M) + \gamma(x, F_M)} \tag{5}$$

where $\gamma(x, D)$ is a distance between a point $x$ and a subspace $D$. $\varphi_\varepsilon(x)$ is zero when $x \in R - G_M$, while one when $x \in F_M$. Because $\gamma(x, F_M)$ and $\gamma(x, R - G_M)$ are continuous and the sum of the two functions keeps at nonzero, $\varphi_\varepsilon(x)$ is continuous. In $\mathcal{L}_v^1(X)$, a linear functional sequence, $f_n(x) = (x, x_n), n = 1, 2, \dots$, can be found. If $\lim_{n \to \infty} f_n(x_1) = f(x_1) = 1$ where $x_1 \in F_M$ is an element of a dense everywhere countable set in $F_M$, based on linearity of the linear functional $\alpha x_1$ can always be found and $\alpha$ is an arbitrary number, which makes $f(\alpha x_1) = \alpha$, namely $f(x) = \frac{x}{x_1}$. It is less difficult to know that $\int |f(x) - \varphi_\varepsilon(x)| dv = \int \left| \frac{x}{x_1} - \varphi_\varepsilon(x) \right| dv$ has not upper bound. If $\lim_{n \to \infty} f_n(x_1) = f(x_1) = 0$ and $x_1 \in R - G_M$, based on linearity of the linear functional $\alpha x_1$ can always be found, which makes $f(\alpha x_1) = 0$, namely $f(x) = 0$. Evidently though, $\int |f(x) - \varphi_\varepsilon(x)| dv = \int |0 - \varphi_\varepsilon(x)| dv = v(F_M)$.

So it's impossible to find a $N$, which makes $\int |f_n(x) - \varphi_\varepsilon(x)| dv < \varepsilon$ for any $n > N$.

That is to say, continuous linear functional set of the separable Hilbert space $X$ is nowhere dense in $\mathcal{L}_v^1(X)$. Evidenced by the same token, the continuous linear functional set of the separable Hilbert space $X$ is nowhere dense in $\mathcal{L}_v^2(X)$ too. Lemma 2 can be obtained.

**Lemma 2.** Continuous linear functional set of the separable Hilbert space $X$ is nowhere dense in $\mathcal{L}_v^2(X)$.

It can be known from lemma 2 that in $\mathcal{L}_v^2(X)$ of a separable Hilbert space the prediction model could not infinitely approximate the optimal interpretation function. In other words, there is an equilibrium problem between the two models. Meanwhile, from Lemma 2, it is less difficult to know that Lemma 3 is true too.

**Lemma 3.** Continuous nonlinear functional set of the separable Hilbert space $X$ is everywhere dense in $\mathcal{L}_v^2(X)$.

When the separable Hilbert space is $\mathcal{L}_v^2(X)$, continue linear functional set on the space is the reproducing kernel Hilbert spaces $\mathcal{H}_K$. When $f_\rho(x)$ usually approximate a linear functional in $\mathcal{H}_K$ while $P(x)$ don't approximate any linear functional or don't exist in $\mathcal{H}_K$, learning and training process of traditional kernel machine learning algorithm can make a linear functional approximate $P(x)$ according Lemma 2. From Lemma 3, the approximate will not be true until $P(x)$ approximate a linear functional. But that is impossible. So, under the situation, the sufficient and necessary conditions for the consistent convergence is $\mathcal{E}^P(f_\rho) = 0$. In other words, when the necessary and sufficient condition is not met, a contradiction between the expected risk and the interpretation distance exists.

## 4. Error estimate of Hypothesis Space

The contradiction between the expected risk and the interpretation distance can be solved by minimizing the sum of the both.

Suppose the optimal solution $f_\mathcal{H}^P(x)$ of the equilibrium problem can be found in the convex subset $\mathcal{H}$ of $\mathcal{L}_\rho^2(X)$ when the marginal probability measure on is $\rho_x$. The deviation between $f \in \mathcal{H}$ and $f_\mathcal{H}^P(x)$ is defined as an error $\mathcal{E}_\mathcal{H}(f) = \mathcal{E}(f) - \mathcal{E}(f_\mathcal{H}^P) + \mathcal{E}^P(f) - \mathcal{E}^P(f_\mathcal{H}^P)$, where $\mathcal{E}(f)$ is an error between $f(x)$ and the real output $y$. If $f: X \to Y$, $\mathcal{E}(f) = \mathcal{E}_\rho(f) = \int_Z (f(x) - y)^2$.

For any function $f \in \mathcal{H}$, $\mathcal{E}_\mathcal{H}(f) \geq 0$ and $\mathcal{E}_\mathcal{H}(f_\mathcal{H}^P) = 0$. Let us focus on that

$$\mathcal{E}(f_{\mu z}) + \mathcal{E}^P(f_{\mu z}) = \mathcal{E}_\mathcal{H}(f_{\mu z}) + \mathcal{E}(f_\mathcal{H}^P) + \mathcal{E}^P(f_\mathcal{H}^P) \tag{12}$$

where $\mathcal{E}_\mathcal{H}(f_{\mu z})$ is a distance between $f_{\mu z}(x)$ and $f_\mathcal{H}^P(x)$, denoted by sample error. $\mathcal{E}(f_\mathcal{H}^P)$ is a distance between $f_\mathcal{H}^P(x)$ and $y$, and $\mathcal{E}^P(f_\mathcal{H}^P)$ is a distance between $f_\mathcal{H}^P(x)$ and $P(x)$, the sum of the two distances is approximate error.

### 4.1 Sample error estimation

From the above formula(12), it can be seen that

$$\mathcal{E}_\mathcal{H}(f) = \mathcal{E}(f) - \mathcal{E}(f_\mathcal{H}^P) + \mathcal{E}^P(f) - \mathcal{E}^P(f_\mathcal{H}^P) \tag{13}$$

The formula can be divided into two parts: $\mathcal{E}(f) - \mathcal{E}(f_\mathcal{H}^P)$ and $\mathcal{E}^P(f) - \mathcal{E}^P(f_\mathcal{H}^P)$.

Probability bound of the former, $\mathcal{E}(f) - \mathcal{E}(f_\mathcal{H}^P)$, can be deduced by Theorem B and Theorem C of reference [1].

**Theorem 1**. Suppose $\mathcal{H}$ is a compact subset of $\mathcal{L}_\rho^2(X)$, and for all $f \in \mathcal{H}$, $|f(x) - y| \leq M$ is true almost everywhere. If

$$\sigma^2 = \sigma^2(\mathcal{H}) = \sup_{f \in \mathcal{H}} \sigma^2(f_Y^2)$$

where $\sigma^2(f_Y^2)$ is a variance of $f_Y^2 = (f(x) - y)^2$, for all $\varepsilon > 0$,

$$prob_{z \in Z^m}\{|\mathcal{E}(f_z) - \mathcal{E}(f_\mathcal{H}^P)| \leq \varepsilon\} \geq 1 - \mathcal{N}\left(\mathcal{H}, \frac{\varepsilon}{16M}\right) 2e^{-\frac{m\varepsilon^2}{8\left(4\sigma^2 + \frac{1}{3}M^2\varepsilon\right)}} \quad (14)$$

where $\mathcal{N}\left(\mathcal{H}, \frac{\varepsilon}{16M}\right)$ is a covering number on $\mathcal{H}$ in the radius $\frac{\varepsilon}{16M}$.

According to lemma 5 in literature [1], it is easy to deduce Theorem 2 in the convex hypothesis space $\mathcal{H}$.

**Theorem 2.** Suppose $\mathcal{H}$ is a compact convex subset of $\mathcal{L}_\rho^2(X)$ which can be sure that the interpretation distance between $f_\mathcal{H}^P$ and $P(x)$ is as small as possible, and for all $f \in \mathcal{H}$, $|f(x) - y| \leq M$ is true almost everywhere. For all $\varepsilon > 0$,

$$prob_{z \in Z^m}\{|\mathcal{E}(f_z) - \mathcal{E}(f_\mathcal{H}^P)| \leq \varepsilon\} \geq 1 - \mathcal{N}\left(\mathcal{H}, \frac{\varepsilon}{24M}\right) 2e^{-\frac{m\varepsilon}{288M^2}} \quad (15)$$

Probability bound of the latter, $\mathcal{E}^P(f) - \mathcal{E}^P(f_\mathcal{H}^P)$, can be obtained by the following process.

Suppose $\mu_z^P(f) = \frac{1}{m}\sum_{i=1}^m (f(x_i) - P(x_i))$ and $\mathcal{E}_z^P(f) = \frac{1}{m}\sum_{i=1}^m (f(x_i) - P(x_i) - \mu_z^P(f))^2$.

Because $\mathcal{E}^P(f) - \mathcal{E}_z^P(f) = \int_Z (f(x) - P(x) - \mu^P(f))^2 - \frac{1}{m}\sum_{i=1}^m (f(x_i) - P(x_i) - \mu_z^P(f))^2$ where $\mu^P(f)$ and $\mu_z^P(f)$ are changeless, based on theorem A of reference [1] Theorem 3 can be obtained.

**Theorem 3.** If $f: X \to Y$ and $P(x)$ is an interpretation function, when $|f(x) - P(x) - \mu_z^P(f)| \leq M_P$ is true almost everywhere for $M_P > 0$, for all $\varepsilon_P > 0$, inequality

$$prob_{z \in Z^m}\{|\mathcal{E}^P(f) - \mathcal{E}_z^P(f)| \leq \varepsilon_P\} \geq 1 - 2e^{-\frac{m\varepsilon_P^2}{2\left(\sigma_P^2 + \frac{1}{3}M_P\varepsilon_P\right)}} \quad (16)$$

holds, where $\sigma_P^2$ is a variance of $(f(x) - P(x) - \mu_z^P(f))^2$.

Theorem 4 can be derived from theorem 3.

**Theorem 4**. Suppose $\mathcal{H}$ is a compact subset of $\mathcal{L}_\rho^2(X)$, if $P(x)$ is the interpretation function and for all $f \in \mathcal{H}$, $|f(x) - P(x) - \mu^P(f)| \leq M_P$ is true almost everywhere, for all $\varepsilon_P > 0$, inequality

$$prob_{z \in Z^m}\{\sup_{f \in \mathcal{H}} |\mathcal{E}^P(f) - \mathcal{E}_z^P(f)| \leq \varepsilon_P\} \geq 1 - 2\mathcal{N}\left(\mathcal{H}, \frac{\varepsilon_P}{16M_P}\right) e^{-\frac{m\varepsilon_P^2}{4\left(2\sigma_P^2 + \frac{1}{3}M_P\varepsilon_P\right)}} \quad (17)$$

holds, where $\sigma_P^2$ is a maximum variance of $(f(x) - P(x) - \mu_z^P(f))^2$,

$$\sigma_P^2 = \sigma_P^2(\mathcal{H}) = \sup_{f \in \mathcal{H}} \sigma_P^2 \left((f(x) - P(x) - \mu_z^P(f))^2\right)$$

Lemma 5 or Lemma 5* give out a linearly dependent bound with $\varepsilon$ when $\mathcal{H}$ is a compact convex subset of $\mathcal{L}_\rho^2(X)$. Then, according to Lemma 5 or Lemma 5*, the probability bound of sample error is given.

**Lemma 5**. Suppose $\mathcal{H}$ is a compact subset of $\mathcal{L}_\rho^2(X)$, if $\varepsilon, \varepsilon_P > 0$, $0 < \delta < 1$, $prob_{z \in Z^m}\{\sup_{f \in \mathcal{H}} |L_z(f)| \leq \varepsilon\} \geq 1 - \delta$, and $prob_{z \in Z^m}\{\sup_{f \in \mathcal{H}} |L_z^P(f)| \leq \varepsilon_P\} \geq 1 - \delta$,

$$prob_{z \in Z^m}\{\mathcal{E}_{\mathcal{H}}(f_z) + \mathcal{E}_{\mathcal{H}}^P(f_z) \leq 2(\varepsilon + \varepsilon_P)\} \geq (1 - \delta)^2 \quad (18)$$

holds.

**Lemma 5***. Suppose $\mathcal{H}$ is a compact subset of $\mathcal{L}_\rho^2(X)$, if $\varepsilon, \varepsilon_P > 0$, $0 < \delta, \delta^P < 1$, $prob_{z \in Z^m}\{\sup_{f \in \mathcal{H}} |L_z(f)| \leq \varepsilon\} \geq 1 - \delta$, and $prob_{z \in Z^m}\{\sup_{f \in \mathcal{H}} |L_z^P(f)| \leq \varepsilon_P\} \geq 1 - \delta^P$,

$$prob_{z \in Z^m}\{\mathcal{E}_{\mathcal{H}}(f_z) + \mathcal{E}_{\mathcal{H}}^P(f_z) \leq 4\varepsilon\} \geq 1 + \delta\delta^P - \delta - \delta^P \quad (19)$$

holds.

In Lemma 5, if $\varepsilon$ is replaced by $\varepsilon/4$, based on Theorem 1 and 4, we can obtain the following conclusion.

**Theorem 5**. Suppose $\mathcal{H}$ is a compact subset of $\mathcal{L}_\rho^2(X)$, for all $f \in \mathcal{H}$ $P(x)$ is interpretation function, $|f(x) - P(x) - \mu^P(f)| \leq M_P$ and $|f(x) - y| \leq M$ is true almost everywhere. If $\sigma_P^2 = \sigma_P^2(\mathcal{H}) = \sup_{f \in \mathcal{H}} \sigma_P^2 \left((f(x) - P(x) - \mu^P(f))^2\right)$ and $\sigma^2 = \sigma^2(\mathcal{H}) = \sup_{f \in \mathcal{H}} \sigma^2(f_Y^2)$, where $\sigma_P^2 \left((f(x) - P(x) - \mu^P(f))^2\right)$ is a variance of $(f(x_i) - P(x_i) - \mu^P(f))^2$ and $\sigma^2(f_Y^2)$ is a variance of $f_Y^2$, for all $\varepsilon > 0$,

$$prob_{z\in Z^m}\{|\mathcal{E}^P(f_z) - \mathcal{E}^P(f_{\mathcal{H}}^P)| + |\mathcal{E}(f_z) - \mathcal{E}(f_{\mathcal{H}}^P)| \leq \varepsilon\} \geq$$

$$\left[1 - 2\mathcal{N}\left(\mathcal{H}, \frac{\varepsilon}{64M_P}\right)e^{-\frac{m\varepsilon^2}{16\left(8\sigma_P^2 + \frac{1}{3}M_P^2\varepsilon\right)}}\right]\left[1 - \mathcal{N}\left(\mathcal{H}, \frac{\varepsilon}{64M}\right)2e^{-\frac{m\varepsilon^2}{32\left(16\sigma^2 + \frac{1}{3}M^2\varepsilon\right)}}\right] \quad (20)$$

Under the condition of no noise, for all $f \in \mathcal{L}_\rho^2(X)$, we have $\sigma^2(f_Y^2) = 0$. [1] So, $\sigma^2 = 0$. Similarly, $\sigma_P^2 = 0$. Exponents in Theorem 5 turn into $\frac{3m}{32M^2}$ and $\frac{3m\varepsilon}{16M_P^2}$. Theorem 5 give out a linearly dependent bound with ε. Meanwhile, Theorem 5* can give out the linearly dependent bound with ε, but need not suppose $\sigma_\rho^2 = \sigma_P^2 = 0$.

**Theorem 5\***. Suppose $\mathcal{H}$ is a compact convex subset of $\mathcal{L}_\rho^2(X)$. If to all $f \in \mathcal{H}$, $P(x)$ is an interpretation function, $|f(x_i) - P(x_i) - \mu^P(f)| \leq M_P$, and $|f(x) - y| \leq M$ is true almost everywhere, for all $\varepsilon_P > 0$, inequality

$$prob_{z\in Z^m}\{|\mathcal{E}^P(f_z) - \mathcal{E}^P(f_{\mathcal{H}}^P)| + |\mathcal{E}(f_z) - \mathcal{E}(f_{\mathcal{H}}^P)| \leq \varepsilon\} \geq$$

$$1 - \mathcal{N}\left(\mathcal{H}, \frac{\varepsilon}{8(3M+2M_P)}\right)e^{-\frac{m\varepsilon}{32(M^2+M_P^2)}\left(\frac{M}{3M+2M_P}\right)^2} \quad (21)$$

holds.

### 4.2 Approximation error estimate

Based on the Hilbert-Schmidt theorem, we can get the theorem 6.

**Theorem 6**. Suppose $\mathcal{H}$ is a Hilbert space, $A$ is a strict positive definite self adjoint compact operator.

(1) If $0 < r \leq s, r \in \mathbb{R}$, for all $a \in \mathcal{H}$, let $\mathcal{L} = Id - \Gamma$, $\Gamma(b-p) = \int(b-p)d\rho$, then

$$\min_{b\in\mathcal{H}}(\|b-a\|^2 + \tau\|b - p - \int(b-p)d\rho\|^2 + \gamma\|A^{-s}b\|^2) \leq$$
$$\|(Id + \tau\mathcal{L}^2 + \gamma A^{-2s})^{-1}[\tau\mathcal{L}^2 p - (\tau\mathcal{L}^2 + \gamma A^{-2s})a]\|^2 + \tau\|\mathcal{L}(Id + \tau\mathcal{L}^2 + \gamma A^{-2s})^{-1}[a -$$
$$(1+\gamma A^{-2s})p]\|^2 + (r+s)^{\frac{r+s}{s}}\gamma^{\frac{r}{s}}(s-r)^{-\frac{r+s}{s}}(1+\tau\mathcal{L}^2)^{-\frac{r+s}{s}}\|A^{-r}(a+\tau\mathcal{L}^2 p)\|^2 \quad (22)$$

(2) If $\|A^{-s}b\| \leq R, R > 0$, for all $a \in \mathcal{H}$,

$$\min_{b\in\mathcal{H}}(\|b-a\|^2 + \tau\|b - p - \int(b-p)d\rho\|^2) \leq \|(Id + \tau\mathcal{L}^2 + \gamma A^{-2s})^{-1}[\tau\mathcal{L}^2 p -$$
$$(\tau\mathcal{L}^2 + \gamma A^{-2s})a]\|^2 + \tau\|\mathcal{L}(Id + \tau\mathcal{L}^2 + \gamma A^{-2s})^{-1}[a - (1+\gamma A^{-2s})p]\|^2 \quad (23)$$

where $\gamma \leq (r+s)^{\frac{r+s}{s-r}} R^{-\frac{2s}{s-r}}(s-r)^{-\frac{r+s}{s-r}}(1+\tau\mathcal{L}^2)^{-\frac{r+s}{s-r}}\|A^{-r}(a+\tau\mathcal{L}^2 p)\|^{\frac{2s}{s-r}}$.

In both cases, $b$ is uniquely exists and finite and in the first part, the optimal $b$ is

$$\hat{b} = (Id + \tau \mathcal{L}^2 + \gamma A^{-2s})^{-1}(a + \tau \mathcal{L}^2 p).$$

Now, in a Hilbert space, a general setting is introduced. Suppose $\nu$ is a Borel measure in $X$ and $A: \mathcal{L}_\nu^2(X) \to \mathcal{L}_\nu^2(X)$ is a strict positive definite compact operator, and $\mathbb{E} = \{g \in \mathcal{L}_\nu^2(X) | \|A^{-s}g\|_\nu < \infty\}$ where $\mathcal{L}_\nu^2(X)$ is a squared integrable function space with Lebesgue measure $\nu$ induced by a quotient space $\mathbb{R}^n$ on $X$. In $\mathbb{E}$, an inner product is defined as $\langle g, h \rangle_\mathbb{E} = \langle A^{-s}g, A^{-s}h \rangle_\nu$. $\mathbb{E}$ is a Hilbert space. So, $A^{-s}: \mathcal{L}_\nu^2(X) \to \mathbb{E}$ is a Hilbert isomorphism. For the general setting, some supposes should be given. $\mathbb{E} \to \mathcal{L}_\nu^2(X)$ can be decomposed into $J_\mathbb{E}: \mathbb{E} \to \mathcal{C}(X)$ and $\mathcal{C}(X) \subset \mathcal{L}_\nu^2(X)$. Suppose $\mathcal{H} = \mathcal{H}_{\mathbb{E},R}$ is $\overline{J_\mathbb{E}(B_R)}$, where $B_R$ is a sphere with radius $R$. If $\mathcal{D}_{\nu\rho}$ is a norm of an operator $J: \mathcal{L}_\nu^2(X) \to \mathcal{L}_\rho^2(X)$, we can obtain Theorem 7.

**Theorem 7**. In the general setting of a Hilbert space, for $0 < r \leq s, r \in \mathbb{R}$, the approximation error

$$\mathcal{E}(f_\mathcal{H}^P) + \mathcal{E}^P(f_\mathcal{H}^P) = \min_{g(x) \in B_R} \left( \|f_\rho(x) - g(x)\|_\rho^2 + \tau \|g(x) - P(x) - \mu^P(g)\|_\rho^2 \right) + \sigma_\rho^2 \leq$$

$$\mathcal{D}_{\nu\rho}^2 \|(Id + \tau \mathcal{L}^2 + \gamma A^{-2s})^{-1}[\tau \mathcal{L}^2 P(x) - (\tau \mathcal{L}^2 + \gamma A^{-2s})f_\rho(x)]\|_\nu^2 + \tau \mathcal{D}_{\nu\rho}^2 \|\mathcal{L}(Id + \tau \mathcal{L}^2 +$$

$$\gamma A^{-2s})^{-1}[f_\rho(x) - (1 + \gamma A^{-2s})P(x)]\|_\nu^2 + \sigma_\rho^2 \tag{24}$$

where

$$\gamma \leq (r+s)^{\frac{r+s}{s-r}} R^{-\frac{2s}{s-r}} (s-r)^{-\frac{r+s}{s-r}} (1 + \tau \mathcal{L}^2)^{-\frac{r+s}{s-r}} \mathcal{D}_{\nu\rho}^2 \left\| A^{-r} \left( f_\rho(x) + \tau \mathcal{L}^2 P(x) \right) \right\|_\nu^{\frac{2s}{s-r}}.$$

**Proof:** We apply Theorem 6 (2). And if $\mathcal{H} = \mathcal{L}_\nu^2(X)$, $a = f_\rho(x)$, and $p = P(x)$, we get

$$\mathcal{E}(f_\mathcal{H}^P) + \mathcal{E}^P(f_\mathcal{H}^P) = \min_{g(x) \in B_R} \left( \|f_\rho(x) - g(x)\|_\rho^2 + \tau \|g(x) - P(x) - \mu^P(g)\|_\rho^2 \right) + \sigma_\rho^2$$

$$\leq \mathcal{D}_{\nu\rho}^2 \min_{g(x) \in B_R} \left( \|f_\rho(x) - g(x)\|_\nu^2 + \tau \|g(x) - P(x) - \mu^P(g)\|_\nu^2 \right) + \sigma_\rho^2$$

$$\leq \mathcal{D}_{\nu\rho}^2 \|(Id + \tau \mathcal{L}^2 + \gamma A^{-2s})^{-1}[\tau \mathcal{L}^2 P(x) - (\tau \mathcal{L}^2 + \gamma A^{-2s})f_\rho(x)]\|_\nu^2$$

$$+ \tau \mathcal{D}_{\nu\rho}^2 \|\mathcal{L}(Id + \tau \mathcal{L}^2 + \gamma A^{-2s})^{-1}[f_\rho(x) - (1 + \gamma A^{-2s})P(x)]\|_\nu^2 + \sigma_\rho^2$$

where

$$\gamma \leq (r+s)^{\frac{r+s}{s-r}} R^{-\frac{2s}{s-r}} (s-r)^{-\frac{r+s}{s-r}} (1 + \tau \mathcal{L}^2)^{-\frac{r+s}{s-r}} \mathcal{D}_{\nu\rho}^2 \left\| A^{-r} \left( f_\rho(x) + \tau \mathcal{L}^2 P(x) \right) \right\|_\nu^{\frac{2s}{s-r}}.$$

□

If $\nu = \rho$, $\mathcal{D}_{\nu\rho} = 1$.

## 4.3 Approximation error estimate in Sobolev space and RKHS

In the section, suppose $X \subset \mathbb{R}^n$ is a compact region with smooth boundary.

**Theorem 8.** If $0 < r < s$, $B_R$ is a sphere with radius $R$ in a conjugate space $H(X)$ on $X$, and $\mathcal{H} = \overline{J_{H(X)}(B_R)}$, the approximation error is

$$\mathcal{E}(f_{\mathcal{H}}^P) + \mathcal{E}^P(f_{\mathcal{H}}^P) \leq \mathcal{D}_{\nu\rho}^2 \left\| (Id + \tau \mathcal{L}^2 + \gamma A^{-2s})^{-1} [\tau \mathcal{L}^2 P(x) - (\tau \mathcal{L}^2 + \gamma A^{-2s}) f_\rho(x)] \right\|_\nu^2 +$$

$$\tau \mathcal{D}_{\nu\rho}^2 \left\| \mathcal{L}(Id + \tau \mathcal{L}^2 + \gamma A^{-2s})^{-1} [f_\rho(x) - (1 + \gamma A^{-2s}) P(x)] \right\|_\nu^2 + \sigma_\rho^2 \qquad (25)$$

where $\gamma \leq (r+s)^{\frac{r+s}{s-r}} (RC)^{-\frac{2s}{s-r}} (s-r)^{-\frac{r+s}{s-r}} (1 + \tau \mathcal{L}^2)^{-\frac{r+s}{s-r}} \mathcal{D}_{\nu\rho}^2 \left\| f_\rho(x) + \tau \mathcal{L}^2 P(x) \right\|_\nu^{\frac{2s}{s-r}}$,

and $C$ is a constant only depends on $r, X$.

**Proof:** Suppose $\Delta: H^2(X) \to \mathcal{L}_\nu^2(X)$ is a Laplacian operator, and $A = (-\Delta + Id)^{-1/2}$. For all $\eta \geq 0$, $A^\eta: \mathcal{L}_\nu^2(X) \to H^\eta(X)$ is a compact linear mapping with bounded inverse. If $C_0, C_1 > 0$, for all $g \in H^\eta(X)$,

$$C_0 \|g\|_\nu \leq \|A^{-\eta} g\|_\nu \leq C_1 \|g\|_\nu \qquad (26)$$

Because $H(X) \subset \mathcal{L}_\nu^2(X)$, we can think $A: \mathcal{L}_\nu^2(X) \to \mathcal{L}_\nu^2(X)$. Then the general setting of a Hilbert space is considered.

If $\mathbb{E}$ is a space in the general setting, a sphere $B_{RC_0}(\mathbb{E})$ with radius $RC_0$ in $\mathbb{E}$ is contained in a sphere $B_R(H(X))$ in $H(X)$. Then

$$\mathcal{E}(f_{\mathcal{H}}^P) + \mathcal{E}^P(f_{\mathcal{H}}^P) = \min_{g(x) \in B_R(H(X))} \left( \|f_\rho(x) - g(x)\|_\rho^2 + \tau \|g(x) - P(x) - \mu^P(g)\|_\rho^2 \right) +$$

$$\sigma_\rho^2 = \min_{g(x) \in B_{RC_0}(\mathbb{E})} \left( \|f_\rho(x) - g(x)\|_\rho^2 + \tau \|g(x) - P(x) - \mu^P(g)\|_\rho^2 \right) + \sigma_\rho^2 \qquad (27)$$

From Theorem 7, we can obtain

$$\min_{g(x) \in B_{RC_0}(\mathbb{E})} \left( \|f_\rho(x) - g(x)\|_\rho^2 + \tau \|g(x) - P(x) - \mu^P(g)\|_\rho^2 \right) + \sigma_\rho^2 \leq \mathcal{D}_{\nu\rho}^2 \left\| (Id + \right.$$

$$\left. \tau \mathcal{L}^2 + \gamma A^{-2s})^{-1} [\tau \mathcal{L}^2 P(x) - (\tau \mathcal{L}^2 + \gamma A^{-2s}) f_\rho(x)] \right\|_\nu^2 +$$

$$\tau \mathcal{D}_{\nu\rho}^2 \left\| \mathcal{L}(Id + \tau \mathcal{L}^2 + \gamma A^{-2s})^{-1} [f_\rho(x) - (1 + \gamma A^{-2s}) P(x)] \right\|_\nu^2 + \sigma_\rho^2 \qquad (28)$$

where $\gamma \leq (r+s)^{\frac{r+s}{s-r}} (RC_0)^{-\frac{2s}{s-r}} (s-r)^{-\frac{r+s}{s-r}} (1 + \tau \mathcal{L}^2)^{-\frac{r+s}{s-r}} \mathcal{D}_{\nu\rho}^2 \left\| A^{-r} \left( f_\rho(x) + \tau \mathcal{L}^2 P(x) \right) \right\|_\nu^{\frac{2s}{s-r}}$.

finally, Eq.(26) is applied, and if $\eta = r$, we can get

$$\left\|A^{-r}\left(f_\rho(x) + \tau\mathcal{L}^2 P(x)\right)\right\|_\nu \leq C_1\|f_\rho(x) + \tau\mathcal{L}^2 P(x)\|_\nu \tag{29}$$

If $C = C_0^{-1}C_1$, $\gamma \leq (r+s)^{\frac{r+s}{s-r}}(RC)^{-\frac{2s}{s-r}}(s-r)^{-\frac{r+s}{s-r}}(1+\tau\mathcal{L}^2)^{-\frac{r+s}{s-r}}\mathcal{D}_{\nu\rho}^2\|f_\rho(x)+\tau\mathcal{L}^2 P(x)\|_\nu^{\frac{2s}{s-r}}$

holds. □

In a reproducing kernel Hilbert space(RKHS), we can get Theorem 9.

**Theorem 9.** Suppose $K$ is a Mercer kernel function, $\nu$ is a Borel measure on $X$, $R > 0$, $\mathcal{H} = \overline{L_K(B_R)}$. For $0 < r < s$, the approximation error is

$$\mathcal{E}(f_\mathcal{H}^P) + \mathcal{E}^P(f_\mathcal{H}^P) \leq \mathcal{D}_{\nu\rho}^2\left\|(Id + \tau\mathcal{L}^2 + \gamma L_K^1)^{-1}[\tau\mathcal{L}^2 P(x) - (\tau\mathcal{L}^2 + \gamma L_K^1)f_\rho(x)]\right\|_\nu^2 +$$

$$\tau\mathcal{D}_{\nu\rho}^2\left\|\mathcal{L}(Id + \tau\mathcal{L}^2 + \gamma L_K^1)^{-1}[f_\rho(x) - (1+\gamma L_K^1)P(x)]\right\|_\nu^2 + \sigma_\rho^2 \tag{30}$$

where $\gamma \leq (r+s)^{\frac{r+s}{s-r}}R^{-\frac{2s}{s-r}}(s-r)^{-\frac{r+s}{s-r}}(1+\tau\mathcal{L}^2)^{-\frac{r+s}{s-r}}\mathcal{D}_{\nu\rho}^2\left\|L_K^{-r/2}\left(f_\rho(x)+\tau\mathcal{L}^2 P(x)\right)\right\|_\nu^{\frac{2s}{s-r}}$.

**Proof:** In Theorem 7, suppose $A = L_K^{1/2}$. Then, from Mercer theorem, for all $f \in \mathcal{L}_\nu^2(X)$, according the general setting, kernel norm $\|f\|_K = \|A^{-1}f\|_\nu$. Theorem 7 is applied. Then if $s = 1$,

$$\mathcal{E}(f_\mathcal{H}^P) + \mathcal{E}^P(f_\mathcal{H}^P) \leq \mathcal{D}_{\nu\rho}^2\left\|(Id + \tau\mathcal{L}^2 + \gamma L_K^{-1})^{-1}[\tau\mathcal{L}^2 P(x) - (\tau\mathcal{L}^2 + \gamma L_K^{-1})f_\rho(x)]\right\|_\nu^2 +$$

$$\tau\mathcal{D}_{\nu\rho}^2\left\|\mathcal{L}(Id + \tau\mathcal{L}^2 + \gamma L_K^{-1})^{-1}[f_\rho(x) - (1+\gamma L_K^{-1})P(x)]\right\|_\nu^2 + \sigma_\rho^2 \tag{31}$$

where $\gamma \leq (r+s)^{\frac{r+s}{s-r}}R^{-\frac{2s}{s-r}}(s-r)^{-\frac{r+s}{s-r}}(1+\tau\mathcal{L}^2)^{-\frac{r+s}{s-r}}\mathcal{D}_{\nu\rho}^2\left\|L_K^{-r/2}\left(f_\rho(x)+\tau\mathcal{L}^2 P(x)\right)\right\|_\nu^{\frac{2s}{s-r}}$.

□

## 5. Optimal error bound of the equilibrium problem

According the general setting in section 4.2, suppose sample size is $m$ and the confidence is $1-\delta, 0 < \delta < 1$. For every $R > 0$, hypothesis space $\mathcal{H} = \mathcal{H}_{\mathbb{E},R}$. We consider $f_\mathcal{H}^P$ and $f_z, z \in Z^m$. In the general setting, the optimal error bound of the equilibrium problem can be found.

**Theorem 10.** For all $m \in \mathbb{N}$, and $r \in \mathbb{R}, 0 < r < s$, in the general setting and the confidence $1-\delta$, the optimal error bound of the equilibrium problem can be found.

$$\alpha(R, M, M_P) + \varepsilon(R, M, M_P) \leq \mathcal{D}_{\nu\rho}^2 M^2 + \tau\mathcal{D}_{\nu\rho}^2 M_P^2 + (3M + 2M_P)^2\left(1 + \frac{M_P^2}{M^2}\right)\vartheta^*(m, \delta) + \sigma_\rho^2.$$

where $\vartheta^*(m,\delta)$ is an optimal solution of $\vartheta = \frac{\varepsilon M^2}{(3M+2M_P)^2(M^2+M_P^2)}$, and $|f(x)-y| \leq M$ is true almost everywhere and $|f(x)-P(x)-\mu^P(f)| \leq M_P$ is true almost everywhere too.

**Proof**: we know that $\mathcal{E}(f_z) + \mathcal{E}^P(f_z) = \mathcal{E}_{\mathcal{H}}(f_z) + \mathcal{E}(f_{\mathcal{H}}^P) + \mathcal{E}^P(f_{\mathcal{H}}^P)$. For $0 < r < s$, Theorem 7 provides probability bound of the approximation error,

$$\mathcal{E}(f_{\mathcal{H}}^P) + \mathcal{E}^P(f_{\mathcal{H}}^P) \leq \mathcal{D}_{\nu\rho}^2 \left\| (Id + \tau\mathcal{L}^2 + \gamma A^{-2s})^{-1} [\tau\mathcal{L}^2 P(x) - (\tau\mathcal{L}^2 + \gamma A^{-2s}) f_\rho(x)] \right\|_\nu^2 +$$

$$\tau \mathcal{D}_{\nu\rho}^2 \left\| \mathcal{L}(Id + \tau\mathcal{L}^2 + \gamma A^{-2s})^{-1} [f_\rho(x) - (1 + \gamma A^{-2s}) P(x)] \right\|_\nu^2 + \sigma_\rho^2 \tag{32}$$

where $\gamma \leq (r+s)^{\frac{r+s}{s-r}} R^{-\frac{2s}{s-r}} (s-r)^{-\frac{r+s}{s-r}} (1+\tau\mathcal{L}^2)^{-\frac{r+s}{s-r}} \mathcal{D}_{\nu\rho}^2 \left\| A^{-r} \left( f_\rho(x) + \tau\mathcal{L}^2 P(x) \right) \right\|_\nu^{\frac{2s}{s-r}}$.

For obtaining the probability bound of the sample error, given $M = M(R) = \|J_{\mathbb{E}}\|R + M_\rho + \|f_\rho\|_\infty$,

$$|f(x)-y| \leq |f(x)| + |y| \leq |f(x)| + |y-f_\rho(x)| + |f_\rho(x)| \leq \|J_{\mathbb{E}}\|R + M_\rho + \|f_\rho\|_\infty \tag{33}$$

Therefore, $|f(x)-y| \leq M$ is true almost everywhere. So similarly, $|f(x) - P(x) - \mu^P(f)| \leq M_P$ is true almost everywhere too. From Theorem 5* and Birman and Solomyak's work[25], the sample error $\varepsilon > 0$ in confidence $1-\delta$ meets

$$\mathcal{N}\left(\mathcal{H}, \frac{\varepsilon}{8(3M+2M_P)}\right) e^{-\frac{m\varepsilon}{32(M^2+M_P^2)} \left(\frac{M}{3M+2M_P}\right)^2} \geq \delta \tag{34}$$

And according to Proposition 6 of section 1.6 in reference [1] we get

$$\frac{m\varepsilon M^2}{32(3M+2M_P)^2(M^2+M_P^2)} + \ln\frac{1}{\delta} - \left(\frac{8(3M+2M_P)RC_{\mathbb{E}}}{\varepsilon}\right)^{1/\ell_{\mathbb{E}}} \leq 0 \tag{35}$$

Suppose $R\|J_{\mathbb{E}}\| \leq \frac{(3M+2M_P)(M^2+M_P^2)}{M^2}$, we can obtain

$$\frac{m\varepsilon^2}{32(3M+2M_P)^2(M^2+M_P^2)} + \ln\frac{1}{\delta} - \left(\frac{8(3M+2M_P)^2(M^2+M_P^2)C_{\mathbb{E}}}{\varepsilon M^2 \|J_{\mathbb{E}}\|}\right)^{1/\ell_{\mathbb{E}}} \leq 0 \tag{36}$$

If $\vartheta = \frac{\varepsilon M^2}{(3M+2M_P)^2(M^2+M_P^2)}$, then

$$\frac{m}{32}\vartheta + \ln\frac{1}{\delta} - \left(\frac{\|J_{\mathbb{E}}\|\vartheta}{8C_{\mathbb{E}}}\right)^{-1/\ell_{\mathbb{E}}} \leq 0 \tag{37}$$

where $\ell_\mathbb{E} > \frac{1}{2}$, in sobolev space $H^s(X)$ which is related to $n$ and $s$, and $s > n/2, \ell_\mathbb{E} = \frac{s}{n} > \frac{1}{2}$.

If $c_0 = \frac{m}{32}$, $c_1 = \ln\frac{1}{\delta}$, $c_2 = \left(\frac{8C_\mathbb{E}}{\|J_\mathbb{E}\|}\right)^{1/\ell_\mathbb{E}}$, $d = 1/\ell_\mathbb{E}$, then we can obtain

$$c_0 \vartheta + c_1 - c_2 \vartheta^{-d} \leq 0 \tag{38}$$

The first derivative of the left-hand side of the above formula is $c_0 + dc_2 \vartheta^{-d-1}$. The second derivative is $-d(d+1)c_2 \vartheta^{-d-2}$. Therefore, $c_0 \vartheta + c_1 - c_2 \vartheta^{-d}$ is a monotonic increase concave function. In a word, if $c_0 + dc_2 \vartheta^{-d-1} = 0$, a solution $\vartheta^*(m, \delta)$ about $\vartheta$ can be found. And

$$\varepsilon(R, M, M_P) = (3M + 2M_P)^2 \left(1 + \frac{M_P^2}{M^2}\right) \vartheta^*(m, \delta) \tag{39}$$

is the optimal bound of sample error from Theorem 5*.

From Theorem 7, we can obtain

$$\alpha(R, M, M_P) \leq \mathcal{D}_{\nu\rho}^2 \left\|(Id + \tau\mathcal{L}^2 + \gamma A^{-2s})^{-1}[\tau\mathcal{L}^2 P(x) - (\tau\mathcal{L}^2 + \gamma A^{-2s})f_\rho(x)]\right\|_\nu^2 +$$

$$\tau\mathcal{D}_{\nu\rho}^2 \left\|\mathcal{L}(Id + \tau\mathcal{L}^2 + \gamma A^{-2s})^{-1}[f_\rho(x) - (1 + \gamma A^{-2s})P(x)]\right\|_\nu^2 + \sigma_\rho^2 \tag{40}$$

where $\gamma \leq (r+s)^{\frac{r+s}{s-r}} R^{-\frac{2s}{s-r}} (s-r)^{-\frac{r+s}{s-r}} (1+\tau\mathcal{L}^2)^{-\frac{r+s}{s-r}} \mathcal{D}_{\nu\rho}^2 \left\|A^{-r}\left(f_\rho(x) + \tau\mathcal{L}^2 P(x)\right)\right\|_\nu^{\frac{2s}{s-r}}$.

It can be seen from the proof of Theorem 6 that

$$\left\|(Id + \tau\mathcal{L}^2 + \gamma A^{-2s})^{-1}[\tau\mathcal{L}^2 P(x) - (\tau\mathcal{L}^2 + \gamma A^{-2s})f_\rho(x)]\right\|_\nu^2 \leq M^2 \tag{41}$$

$$\left\|\mathcal{L}(Id + \tau\mathcal{L}^2 + \gamma A^{-2s})^{-1}[f_\rho(x) - (1 + \gamma A^{-2s})P(x)]\right\|_\nu^2 \leq M_P^2 \tag{42}$$

We can get $\alpha(R, M, M_P) \leq \mathcal{D}_{\nu\rho}^2 M^2 + \tau\mathcal{D}_{\nu\rho}^2 M_P^2 + \sigma_\rho^2$. And

$$\alpha(R, M, M_P) + \varepsilon(R, M, M_P) \leq$$

$$\mathcal{D}_{\nu\rho}^2 M^2 + \tau\mathcal{D}_{\nu\rho}^2 M_P^2 + (3M + 2M_P)^2 \left(1 + \frac{M_P^2}{M^2}\right) \vartheta^*(m, \delta) + \sigma_\rho^2. \tag{43}$$

Then, $R\|J_\mathbb{E}\| \leq \frac{(3M+2M_P)(M^2+M_P^2)}{M^2}$ will be proved. Because $\|J_\mathbb{E}\|R + M_\rho + \|f_\rho\|_\infty = M$, then

$$(3M + 2M_P)(M^2 + M_P^2) - M^2(M - M_\rho - \|f_\rho\|_\infty) = 2M^3 + (2M_P + M_\rho +$$

$$\|f_\rho\|_\infty)M^2 + 3M_P^2 M + 2M_P^3 > 0 \qquad (44)$$

□

According to Formula (43), if $M$ and $M_P$ go down simultaneous, the sample error and the approximation error will also be reduced. But according subsection 3.2, there are contradiction between $M$ and $M_P$. One of them goes down, which is going to cause another increase.

In the next section, we will prove that the optimal $R$, $M$ and $M_P$, can be found in a new learning framework for improving the interpretability of the predication model without degradation of the generalization performance in some conditions.

## 6. Learning framework and Conditions for global optimal solution

Based on Tihonov regularized learning framework[23-24] and the evaluation formula of the interpretability, a learning framework for improving the interpretability of the prediction model can be obtained.

$$f_{z,\lambda} = \mathrm{argmin}_{f \in \mathcal{H}_K} \left\{ \frac{1}{m}\sum_{i=1}^m (f(x_i) - y_i)^2 + \lambda \|f\|_K^2 + \frac{1}{m}\sum_{i=1}^m \left(f(x_i) - P(x_i) - \frac{1}{m}\sum_{i=1}^m |f(x_i) - P(x_i)|\right)^2 \right\} \qquad (6)$$

Now let us prove the optimal problem is solvable and unique, and find the sufficient condition for a global optimal solution.

In a close subspace $\mathcal{H}$ of $\mathcal{L}_\rho^2(X)$, $f_\mathcal{H}^P(x)$ is defined as a optimal function which has the distance as small as possible with $f_\rho(x)$, while has the smallest distance with the interpretation model $P(x)$. We will prove that if the close subspace $\mathcal{H}$ is convex, $f_\mathcal{H}^P(x)$ must be solvable and unique.

**Lemma 4.** A compact space $\mathcal{H}$ is a convex subset of $\mathcal{L}_\rho^2(X)$, then Eq. (6) must have unique solution $f_\mathcal{H}^P(x) \in \mathcal{H}$.

**Proof:** In $\mathcal{H}$, suppose $s = \overline{f_\mathcal{H}^P f}$ is a segment with two endpoints $f_\mathcal{H}^P$ and $f$. Because $\mathcal{H}$ is a convex subset, and $s \subset \mathcal{H}$, and in $\mathcal{L}_\rho^2(X)$ the sum of the distance between $f_\mathcal{H}^P(x)$ and $f_\rho(x)$ and the distance between $f_\mathcal{H}^P(x)$ and $P(x)$ is the smallest, all $f \in s$,

$$\int_Z \left(f_{\mathcal{H}}^P(x) - P(x) - \mu(f_{\mathcal{H}}^P, P)\right)^2 + \int_Z \left(f_{\mathcal{H}}^P(x) - f_\rho(x)\right)^2 \leq \int_Z \left(f(x) - P(x) - \mu(f,P)\right)^2 + \int_Z \left(f(x) - f_\rho(x)\right)^2. \tag{7}$$

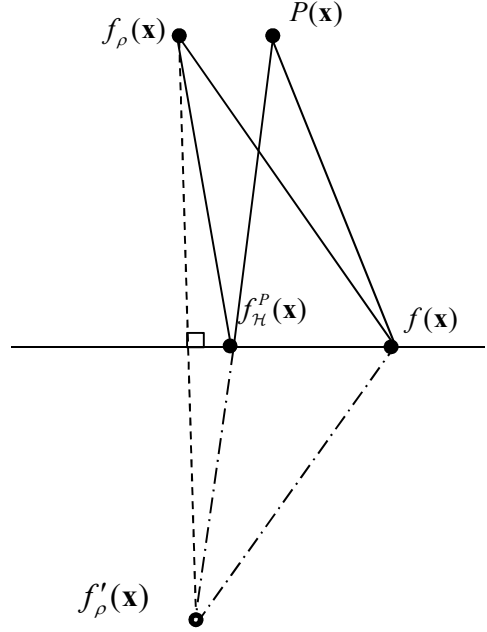

Fig.2 The unique solution of the new learning problem (I)

It can be known from the triangle cosine theorem that in $\triangle P f_{\mathcal{H}}^P f$

$$\int_Z \left[\left(f(x) - P(x) - \mu(f,P)\right)^2 - \left(f_{\mathcal{H}}^P(x) - P(x) - \mu(f_{\mathcal{H}}^P, P)\right)^2\right] = \int_Z \left(f_{\mathcal{H}}^P(x) - f(x) - \mu(f_{\mathcal{H}}^P, f)\right)^2 - 2\int_Z |f_{\mathcal{H}}^P(x) - f(x) - \mu(f_{\mathcal{H}}^P, f)||f_{\mathcal{H}}^P(x) - P(x) - \mu(f_{\mathcal{H}}^P, P)|cos\angle P f_{\mathcal{H}}^P f \tag{8}$$

And in $\triangle f_\rho f_{\mathcal{H}}^P f$,

$$\int_Z \left[\left(f(x) - f_\rho(x)\right)^2 - \left(f_{\mathcal{H}}^P(x) - f_\rho(x)\right)^2\right] = \int_Z \left[\left(f_{\mathcal{H}}^P(x) - f(x)\right)^2 - 2|f_{\mathcal{H}}^P(x) - f(x)||f_{\mathcal{H}}^P(x) - f_\rho(x)|cos\angle f_\rho f_{\mathcal{H}}^P f\right]. \tag{9}$$

Combining the above two equations, we can get

$$\mathcal{E}_{\mathcal{H}}^P(f) = \int_Z \left(f_{\mathcal{H}}^P(x) - f(x) - \mu(f_{\mathcal{H}}^P, f)\right)^2 + \int_Z \left(f_{\mathcal{H}}^P(x) - f(x)\right)^2 - 2\left(\int_Z |f_{\mathcal{H}}^P(x) - f(x) - \mu(f_{\mathcal{H}}^P, f)||f_{\mathcal{H}}^P(x) - P(x) - \mu(f_{\mathcal{H}}^P, P)|cos\angle P f_{\mathcal{H}}^P f + \int_Z |f_{\mathcal{H}}^P(x) - f(x)||f_{\mathcal{H}}^P(x) - f_\rho(x)|cos\angle f_\rho f_{\mathcal{H}}^P f\right). \tag{10}$$

Suppose $f(x) = f'^P_{\mathcal{H}}(x)$, and $f'^P_{\mathcal{H}}(x)$ also is the optimal one, then

$$\int_Z \left|f^P_{\mathcal{H}}(x) - f'^P_{\mathcal{H}}(x) - \mu\left(f^P_{\mathcal{H}}, f'^P_{\mathcal{H}}\right)\right|\left(\left|f^P_{\mathcal{H}}(x) - f'^P_{\mathcal{H}}(x) - \mu\left(f^P_{\mathcal{H}}, f'^P_{\mathcal{H}}\right)\right| - 2\left|f^P_{\mathcal{H}}(x) - P(x) - \mu(f^P_{\mathcal{H}}, P)\right|\cos\angle Pf^P_{\mathcal{H}}f'^P_{\mathcal{H}}\right) + \int_Z \left|f^P_{\mathcal{H}}(x) - f'^P_{\mathcal{H}}(x)\right|\left(\left|f^P_{\mathcal{H}}(x) - f'^P_{\mathcal{H}}(x)\right| - 2\left|f^P_{\mathcal{H}}(x) - f_\rho(x)\right|\cos\angle f_\rho f^P_{\mathcal{H}} f'^P_{\mathcal{H}}\right) = 0. \tag{11}$$

Because $\mathcal{H}$ is convex, from the above formula, it can be seen that there are a function $f^{*P}_{\mathcal{H}}(x)$ between $f^P_{\mathcal{H}}$ and $f'^P_{\mathcal{H}}$, which can guarantee $\left|f^P_{\mathcal{H}}(x) - f^{*P}_{\mathcal{H}}(x) - \mu(f^P_{\mathcal{H}}, f^{*P}_{\mathcal{H}})\right| < \left|f^P_{\mathcal{H}}(x) - f'^P_{\mathcal{H}}(x) - \mu(f^P_{\mathcal{H}}, f'^P_{\mathcal{H}})\right|$ and $\left|f^P_{\mathcal{H}}(x) - f^{*P}_{\mathcal{H}}(x)\right| < \left|f^P_{\mathcal{H}}(x) - f'^P_{\mathcal{H}}(x)\right|$. $\angle Pf^P_{\mathcal{H}}f^{*P}_{\mathcal{H}} < \angle Pf^P_{\mathcal{H}}f'^P_{\mathcal{H}}$ and $\angle f_\rho f^P_{\mathcal{H}} f^{*P}_{\mathcal{H}} < \angle f_\rho f^P_{\mathcal{H}} f'^P_{\mathcal{H}}$. And $\mathcal{E}^P_{\mathcal{H}}\left(f^{*P}_{\mathcal{H}}(x)\right) < 0$.

If so, we can always find a better function $f^{*P}_{\mathcal{H}}(x)$ than $f^P_{\mathcal{H}}(x)$ and $f'^P_{\mathcal{H}}(x)$, as shown in the following figure. Therefore, the conclusion is a contradiction to the previous hypothesis. So $f^P_{\mathcal{H}}(x)$ is the unique solution that meets the condition in $\mathcal{L}^2_\rho(X)$.

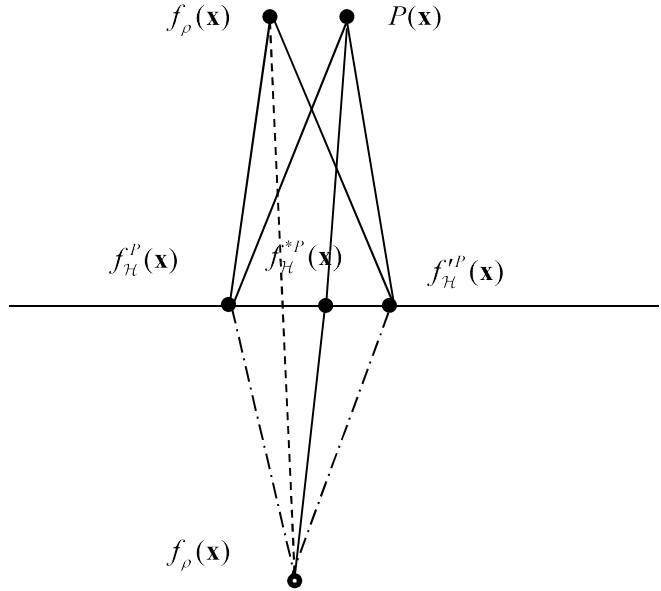

Fig.3 The unique solution of the new learning problem (II)

## 7. Interpretably Least-squares Support Vector Machine(ILSSVM)

A constraint term (105) are added into the least-squares SVM (LSSVM) original regression problem. The new primal problem is

$$\underset{\mathbf{w}\in R^m, b\in R, \mathbf{e}\in R^N, \boldsymbol{\tau}\in R^N}{\operatorname{argmin}} J(\mathbf{w},b,\mathbf{e},\boldsymbol{\tau}) = \frac{1}{2}\mathbf{w}^T\mathbf{w} + \varphi\frac{1}{2}\sum_{k=1}^{N}e_k^2 + \sigma\left(\frac{1}{2}\sum_{k=1}^{N}\tau_k^2\right) \qquad (104)$$

$$s.t. \quad \mathbf{w}\bullet\mathbf{x}_k + b - y_k + e_k = 0, k=1,2,...,N,$$

$$\boldsymbol{\mu}^T\mathbf{C}(\mathbf{x}_k) - y_k - \frac{1}{N}\sum_{i=1}^{N}\left(\boldsymbol{\mu}^T\mathbf{C}(\mathbf{x}_i) - y_i\right) + \tau_k = 0, \qquad (105)$$

$$\mathbf{e} = \left(e_1, e_2, ..., e_N\right)^T, \boldsymbol{\tau} = \left(\tau_1, \tau_2, ..., \tau_N\right)^T,$$

$$k = 1, 2, ..., N, h \geq 0.$$

where $\mathbf{w}\bullet\mathbf{x}_k = \mathbf{w}^T\mathbf{x}_k$.

In the primal problem, to guarantee the predictive accuracy and interpretability of the optimal regression model, a regular term $\sigma\left(\frac{1}{2}\sum_{k=1}^{N}\tau_k^2\right)$ is added to minimize $\tau_k$. $\sigma$ denotes the importance of the interpretability. When the complete relationships between attributes are not correct, $\sigma = 0$, and the primal problem turns into the standard LSSVM original regression problem. $\boldsymbol{\mu}^T\mathbf{C}(\mathbf{x}_k)$ is the interpretability model which is introduced in (Zhao 2018)[14].

## 8. The Proof of Main results

In section, we will prove Theorem 4, Theorem 6, Lemma 5 and Lemma 5*. Firstly, we give the following Lemma 6.

Suppose $f_1, f_2 \in \mathcal{L}_\rho^2(X)$, firstly, for all $z \in Z^m$, $|L_z^P(f_1) - L_z^P(f_2)|$ is estimated, where $L_z^P(f) = \mathcal{E}^P(f) - \mathcal{E}_z^P(f)$ is Lipshitz estimation. Then Lemma 6 can be obtained.

**Lemma 6.** On a completely measurable set $U \subset Z$, if $|f_j(x) - P(x) - \mu_z^P(f_j)| \leq M_P, j = 1,2$, for $z \in U^m$,

$$|L_z^P(f_1) - L_z^P(f_2)| \leq 8M_P\|f_1 - f_2\|_\infty \qquad (47)$$

**Proof**: Because

$$\left(f_1(x) - P(x) - \mu^P(f_1)\right)^2 - \left(f_2(x) - P(x) - \mu^P(f_2)\right)^2 = |f_1(x) - f_2(x) - \mu^P(f_1) +$$

$$\mu^P(f_2)| |\left(f_1(x) - P(x) - \mu^P(f_1)\right) + \left(f_2(x) - P(x) - \mu^P(f_2)\right)|, \tag{48}$$

we get

$$|\mathcal{E}^P(f_1) - \mathcal{E}^P(f_2)| \le 2\|f_1 - f_2\|_\infty \int_Z |\left(f_1(x) - P(x) - \mu^P(f_1)\right) + \left(f_2(x) - P(x) - \mu^P(f_2)\right)| \le 4\|f_1 - f_2\|_\infty M_P. \tag{49}$$

So,

$$|\mathcal{E}_Z^P(f_1) - \mathcal{E}_Z^P(f_2)| = \left|\frac{1}{m}\sum_{i=1}^m \left(f_1(x_i) - P(x_i) - \mu^P(f_1)\right)^2 - \frac{1}{m}\sum_{i=1}^m \left(f_2(x_i) - P(x_i) - \mu^P(f_2)\right)^2\right| \le 4\|f_1 - f_2\|_\infty M_P. \tag{50}$$

And

$$\left|L_z^P(f_1) - L_z^P(f_2)\right| = \left|\mathcal{E}^P(f_1) - \mathcal{E}_z^P(f_1) - \mathcal{E}^P(f_2) + \mathcal{E}_z^P(f_2)\right| \le 8M_P \|f_1 - f_2\|_\infty . \qquad \square$$

If $\mathcal{H} \subseteq \mathcal{L}_\rho^2(X)$, and for all $f \in \mathcal{H}$, $|f(x) - P(x) - \mu^P(f)| \le M_P$ is true almost everywhere, the inequalitys $|\mathcal{E}^P(f_1) - \mathcal{E}^P(f_2)| \le 4M_P\|f_1 - f_2\|_\infty$ and $|\mathcal{E}_z^P(f_1) - \mathcal{E}_z^P(f_2)| \le 4M_P\|f_1 - f_2\|_\infty$ indicate that $\mathcal{E}^P, \mathcal{E}_z^P: \mathcal{H} \to \mathbb{R}$ is continuous.

**Lemma 7.** If $\mathcal{H} = S_1 \cup \cdots \cup S_\ell$ and $\varepsilon > 0$, then

$$\text{prob}_{z \in Z^m}\{\sup_{f \in \mathcal{H}} |L_z^P(f)| \ge \varepsilon_P\} \le \sum_{j=1}^\ell \text{prob}_{z \in Z^m}\{\sup_{f \in S_j} |L_z^P(f)| \ge \varepsilon_P\} \tag{51}$$

**Proof:** Because of the equivalence property

$$\sup_{f \in \mathcal{H}} |L_z^P(f)| \ge \varepsilon_P \Leftrightarrow \exists j \le \ell \text{ s.t. } \sup_{f \in S_j} |L_z^P(f)| \ge \varepsilon_P \tag{52}$$

and a fact, that the union probability of some events is bounded by the sum of the probabilities of these events, the result of the lemma can be obtained. $\square$

**Proof of Theorem 4.** If $\ell_P = \mathcal{N}_P\left(\mathcal{H}, \frac{\varepsilon_P}{8M_P}\right)$, there some disks $D_j, j = 1,2,\ldots,\ell_P$, to cover $\mathcal{H}$, which makes $f_j$ as the center and $\frac{\varepsilon_P}{8M_P}$ as its radius. On $U$ which is a completely measurable set including $x_i$, we have $|f(x_i) - P(x_i) - \mu^P(f)| \le M_P$. From Lemma 6, for all $z \in U^m$ and $f \in D_j$, we have

$$|L_z^P(f_1) - L_z^P(f_2)| \le 8M_P\|f_1 - f_2\|_\infty \le 8M_P \frac{\varepsilon_P}{8M_P} = \varepsilon_P. \tag{53}$$

Then,

$$\sup_{f \in D_j}|L_z^P(f)| \geq 2\varepsilon_P \Longrightarrow |L_z^P(f_j)| \geq \varepsilon_P. \tag{54}$$

Therefore, for $j = 1, \cdots, \ell_P$, from Theorem 3 we can obtain

$$\Prob_{z \in Z^m}\left\{\sup_{f \in D_j}\left|\mathcal{E}^P(f) - \mathcal{E}_z^P(f)\right| \geq 2\varepsilon_P\right\} \leq \prob_{z \in Z^m}\left\{|L_z^P(f_j)| \geq \varepsilon_P\right\} \leq 2e^{-\frac{m\varepsilon_P^2}{2(\sigma_P^2 + \frac{1}{3}M_P\varepsilon_P)}}. \tag{55}$$

Now, we replace $\varepsilon_P$ with $\varepsilon_P/2$, from Lemma 7 we obtain the following conclusion.

$$\Prob_{z \in Z^m}\left\{\sup_{f \in \mathcal{H}}\left|\mathcal{E}^P(f) - \mathcal{E}_z^P(f)\right| \leq \varepsilon_P\right\} \geq 1 - \mathcal{N}\left(\mathcal{H}, \frac{\varepsilon_P}{16M_P}\right)2e^{-\frac{m\varepsilon_P^2}{4(2\sigma_P^2 + \frac{1}{3}M_P^2\varepsilon_P)}} \quad \square$$

**Proof of Lemma 5**. At least the probability $(1 - \delta)^2$, we have

$$|\mathcal{E}(f_z) - \mathcal{E}_z(f_z)| \leq \varepsilon \text{ 和 } |\mathcal{E}^P(f_z) - \mathcal{E}_z^P(f_z)| \leq \varepsilon_P, \tag{56}$$

Therefore,

$$\mathcal{E}(f_z) + \mathcal{E}^P(f_z) \leq \mathcal{E}_z(f_z) + \mathcal{E}_z^P(f_z) + \varepsilon_P + \varepsilon \tag{57}$$

In the same way, we have

$$\mathcal{E}_z(f_z) + \mathcal{E}_z^P(f_z) \leq \mathcal{E}(f_z) + \mathcal{E}^P(f_z) + \varepsilon_P + \varepsilon \tag{58}$$

Moreover, since in $\mathcal{H}$, $f_z$ minimizes $\mathcal{E}_z$ and $\mathcal{E}_z^P$, we have

$$\mathcal{E}_z(f_z) + \mathcal{E}_z^P(f_z) \leq \mathcal{E}_z(f_\mathcal{H}^P) + \mathcal{E}_z^P(f_\mathcal{H}^P) \tag{59}$$

So, in accordance with the probability of at least $(1 - \delta)^2$, we get

$$\mathcal{E}(f_z) + \mathcal{E}^P(f_z) \leq \mathcal{E}_z(f_z) + \mathcal{E}_z^P(f_z) + \varepsilon_P + \varepsilon \leq \mathcal{E}_z(f_\mathcal{H}^P) + \mathcal{E}_z^P(f_\mathcal{H}^P) + \varepsilon_P + \varepsilon \leq \mathcal{E}(f_\mathcal{H}^P) + \mathcal{E}^P(f_\mathcal{H}^P) + 2\varepsilon_P + 2\varepsilon \tag{60}$$

Therefore, $\mathcal{E}_\mathcal{H}(f_z) + \mathcal{E}_\mathcal{H}^P(f_z) \leq 2(\varepsilon_P + \varepsilon)$ $\square$

**Proof of Lemma 5***. At least the probability $(1 - \delta)(1 - \delta^P)$, we have

$$|\mathcal{E}(f_z) - \mathcal{E}_z(f_z)| \leq \varepsilon \text{ 和 } |\mathcal{E}^P(f_z) - \mathcal{E}_z^P(f_z)| \leq \varepsilon, \tag{61}$$

Therefore

$$\mathcal{E}(f_z) + \mathcal{E}^P(f_z) \leq \mathcal{E}_z(f_z) + \mathcal{E}_z^P(f_z) + 2\varepsilon. \tag{62}$$

In the same way, we have

$$\mathcal{E}_{\mathbf{z}}(f_{\mathcal{H}}^{P})+\mathcal{E}_{\mathbf{z}}^{P}(f_{\mathcal{H}}^{P}) \leq \mathcal{E}(f_{\mathcal{H}}^{P})+\mathcal{E}^{P}(f_{\mathcal{H}}^{P})+2\varepsilon. \tag{63}$$

Moreover, since in $\mathcal{H}$ $f_{\mathbf{z}}$ minimizes $\mathcal{E}_{\mathbf{z}}$, we have

$$\mathcal{E}_{\mathbf{z}}(f_{\mathbf{z}})+\mathcal{E}_{\mathbf{z}}^{P}(f_{\mathbf{z}}) \leq \mathcal{E}_{\mathbf{z}}(f_{\mathcal{H}}^{P})+\mathcal{E}_{\mathbf{z}}^{P}(f_{\mathcal{H}}^{P}). \tag{64}$$

So, In accordance with the probability of at least $(1-\delta)(1-\delta^P)$, we get

$$\mathcal{E}(f_{\mathbf{z}})+\mathcal{E}^{P}(f_{\mathbf{z}}) \leq \mathcal{E}_{\mathbf{z}}(f_{\mathbf{z}})+\mathcal{E}_{\mathbf{z}}^{P}(f_{\mathbf{z}})+2\varepsilon \leq \mathcal{E}_{\mathbf{z}}(f_{\mathcal{H}}^{P})+\mathcal{E}_{\mathbf{z}}^{P}(f_{\mathcal{H}}^{P})+2\varepsilon \leq \mathcal{E}(f_{\mathcal{H}}^{P})+\mathcal{E}^{P}(f_{\mathcal{H}}^{P})+4\varepsilon. \tag{65}$$

Therefore, $\mathcal{E}_{\mathcal{H}}(f_{\mathbf{z}}) + \mathcal{E}_{\mathcal{H}}^{P}(f_{\mathbf{z}}) \leq 4\varepsilon$  □

**Proof of Theorem 5\*.** To prove Theorem 5\*, we have to proof Lemma 8.

**Lemma 8.** Given $\mathcal{H}$ is a compact convex subset of $\mathcal{L}_{\rho}^{2}(X)$ which can be sure that the interpretation distance between $f_{\mathcal{H}}^{P}$ and $P(x)$ is as small as possible, then for all $f \in \mathcal{H}$,

$$\int_{Z} \left( \left|f_{\mathcal{H}}^{P}(x) - f(x) - \mu(f, f_{\mathcal{H}}^{P})\right| + \left|f_{\mathcal{H}}^{P}(x) - f(x)\right| \right)^{2} \leq \mathcal{E}_{\mathcal{H}}(f) \tag{66}$$

where $\mathcal{E}_{\mathcal{H}}(f) = \mathcal{E}(f) - \mathcal{E}(f_{\mathcal{H}}^{P}) + \mathcal{E}^{P}(f) - \mathcal{E}^{P}(f_{\mathcal{H}}^{P})$.

**Proof:** Firstly, we consider the first case.

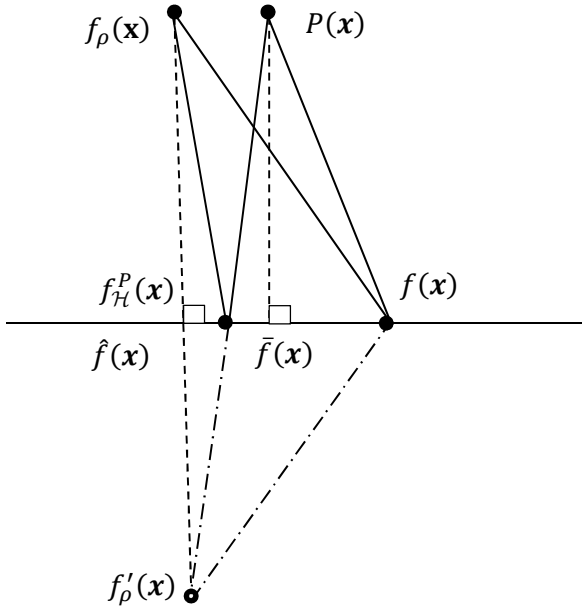

Fig.4 The first case of the compact convex subset $\mathcal{H}$

From the triangle cosine theorem, we have

$$\mathcal{E}^P(f) - \mathcal{E}^P(f_{\mathcal{H}}^P) = \int_Z (f_{\mathcal{H}}^P(\mathbf{x}) - f(\mathbf{x}) - \mu(f, f_{\mathcal{H}}^P))^2 - 2\int_Z |f_{\mathcal{H}}^P(\mathbf{x}) - f(\mathbf{x}) - \mu(f, f_{\mathcal{H}}^P)||f_{\mathcal{H}}^P(\mathbf{x}) - P(\mathbf{x}) - \mu^P(f_{\mathcal{H}}^P)|\cos\angle Pf_{\mathcal{H}}^P f \quad (67)$$

And

$$\mathcal{E}(f) - \mathcal{E}(f_{\mathcal{H}}^P) = \int_Z (f_{\mathcal{H}}^P(\mathbf{x}) - f(\mathbf{x}))^2 - 2\int_Z |f_{\mathcal{H}}^P(\mathbf{x}) - f(\mathbf{x})||f_{\mathcal{H}}^P(\mathbf{x}) - f_\rho(\mathbf{x})|\cos\angle f_\rho f_{\mathcal{H}}^P f \quad (68)$$

From Eq.(67) and Eq.(68), we get

$$\mathcal{E}_{\mathcal{H}}(f) = \mathcal{E}(f) - \mathcal{E}(f_{\mathcal{H}}^P) + \mathcal{E}^P(f) - \mathcal{E}^P(f_{\mathcal{H}}^P)$$
$$= \int_Z (f_{\mathcal{H}}^P(\mathbf{x}) - f(\mathbf{x}))^2 +$$
$$2\int_Z \left(|f_{\mathcal{H}}^P(\mathbf{x}) - f(\mathbf{x})||f_{\mathcal{H}}^P(\mathbf{x}) - \hat{f}(\mathbf{x})| - |f_{\mathcal{H}}^P(\mathbf{x}) - f(\mathbf{x}) - \mu(f, f_{\mathcal{H}}^P)||f_{\mathcal{H}}^P(\mathbf{x}) - \overline{f}(\mathbf{x}) - \mu^-(f_{\mathcal{H}}^P)|\right)$$
$$+ \int_Z (f_{\mathcal{H}}^P(\mathbf{x}) - f(\mathbf{x}) - \mu(f, f_{\mathcal{H}}^P))^2$$

(69)

Suppose $C_1' = |f_{\mathcal{H}}^P(\mathbf{x}) - P(\mathbf{x}) - \mu^P(f_{\mathcal{H}}^P)|\cos\angle Pf_{\mathcal{H}}^P f = |f_{\mathcal{H}}^P(\mathbf{x}) - \overline{f}(\mathbf{x}) - \mu^-(f_{\mathcal{H}}^P)|$ and

$C_2 = |f_{\mathcal{H}}^P(\mathbf{x}) - f_\rho(\mathbf{x})|\cos\angle f_\rho f_{\mathcal{H}}^P f = |f_{\mathcal{H}}^P(\mathbf{x}) - \hat{f}(\mathbf{x})|$, we have

$$\mathcal{E}_{\mathcal{H}}(f) = \int_Z \left(|f_{\mathcal{H}}^P(\mathbf{x}) - f(\mathbf{x})| + |f_{\mathcal{H}}^P(\mathbf{x}) - f(\mathbf{x}) - \mu(f, f_{\mathcal{H}}^P)|\right)^2 +$$
$$2\int_Z \left(|f_{\mathcal{H}}^P(\mathbf{x}) - f(\mathbf{x})|C_2 - |f_{\mathcal{H}}^P(\mathbf{x}) - f(\mathbf{x}) - \mu(f, f_{\mathcal{H}}^P)|C_1'\right) - \quad (70)$$
$$2\int_Z |f_{\mathcal{H}}^P(\mathbf{x}) - f(\mathbf{x})||f_{\mathcal{H}}^P(\mathbf{x}) - f(\mathbf{x}) - \mu(f, f_{\mathcal{H}}^P)|$$

Since $\mathcal{E}_{\mathcal{H}}(f)$ is greater than zero, when $f$ infinite close to $f_{\mathcal{H}}^P$, we have

$$2\int_Z \left(|f_{\mathcal{H}}^P(\mathbf{x}) - f(\mathbf{x})|C_2 - |f_{\mathcal{H}}^P(\mathbf{x}) - f(\mathbf{x}) - \mu(f, f_{\mathcal{H}}^P)|C_1' - |f_{\mathcal{H}}^P(\mathbf{x}) - f(\mathbf{x})||f_{\mathcal{H}}^P(\mathbf{x}) - f(\mathbf{x}) - \mu(f, f_{\mathcal{H}}^P)|\right) \geq 0$$

Therefore,

$$\mathcal{E}_{\mathcal{H}}(f) \geq \int_Z \left(|f_{\mathcal{H}}^P(x) - f(x) - \mu(f, f_{\mathcal{H}}^P)| + |f_{\mathcal{H}}^P(x) - f(x)|\right)^2 \quad (71)$$

*For the second case,*

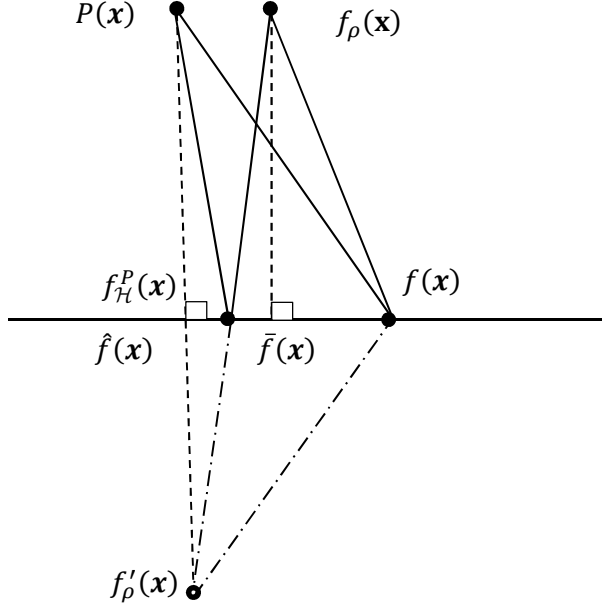

We can easily obtain

$$\mathcal{E}_{\mathcal{H}}(f) \geq \int_Z \left( \left| f_{\mathcal{H}}^P(x) - f(x) - \mu(f, f_{\mathcal{H}}^P) \right| + \left| f_{\mathcal{H}}^P(x) - f(x) \right| \right)^2 \quad (72)$$

The proof of Lemma 8 is completed. □

Let us focus on the function $\ell(f): Z \to Y$, $\ell(f) = (f(x) - y)^2 - (f_{\mathcal{H}}^P(x) - y)^2 + (f(x) - P(x) - \mu^P(f))^2 - (f_{\mathcal{H}}^P(x) - P(x) - \mu^P(f_{\mathcal{H}}^P))^2$. It can be abbreviated to

$$\ell(f) = (f - y)^2 - (f_{\mathcal{H}}^P - y)^2 + (f - P - \mu^P(f))^2 - (f_{\mathcal{H}}^P - P - \mu^P(f_{\mathcal{H}}^P))^2 \quad (73)$$

Then, $E\ell(f) = \mathcal{E}_{\mathcal{H}}(f) = \mathcal{E}(f) - \mathcal{E}(f_{\mathcal{H}}^P) + \mathcal{E}^P(f) - \mathcal{E}^P(f_{\mathcal{H}}^P)$, and for $z \in Z^m$, $E_z \ell(f) = \mathcal{E}_{\mathcal{H},z}(f) = \mathcal{E}_z(f) - \mathcal{E}_z(f_{\mathcal{H}}^P) + \mathcal{E}_z^P(f) - \mathcal{E}_z^P(f_{\mathcal{H}}^P)$. Moreover, we know that for all $f \in \mathcal{H}$, $|\ell(f)(x, y)| \leq M^2 + M_P^2$ is true almost everywhere.

Suppose $\sigma^2 = \sigma^2(\ell(f))$ is a variance of $\ell(f)$. We can obtain Lemma 9.

**Lemma 9.** For all $f \in \mathcal{H}$, $\sigma^2 \leq 4\max(M^2, M_P^2) \mathcal{E}_{\mathcal{H}}(f)$.

**Proof**: Since

$$\sigma^2 \leq E\ell^2(f) = E\left[(f-y)^2 - (f_{\mathcal{H}}^P - y)^2 + (f - P - \mu^P(f))^2 - (f_{\mathcal{H}}^P - P - \mu^P(f_{\mathcal{H}}^P))^2\right]^2 \leq$$

$$E\left[2M(f - f_{\mathcal{H}}^P) + 2M_P\left(f - f_{\mathcal{H}}^P - \mu^P(f) + \mu^P(f_{\mathcal{H}}^P)\right)\right]^2 \leq 4max(M^2, M_P^2)E\left[(f - f_{\mathcal{H}}^P) + \right.$$

$$\left(f - f_{\mathcal{H}}^P - \mu^P(f) + \mu^P(f_{\mathcal{H}}^P)\right)\right]^2 \tag{74}$$

And because $E\left[(f - f_{\mathcal{H}}^P) + (f - f_{\mathcal{H}}^P - \mu(f, f_{\mathcal{H}}^P))\right]^2 \leq \mathcal{E}_{\mathcal{H}}(f)$, Lemma 9 is true. □

**Lemma 10**. If $f \in \mathcal{H}$, for all $\varepsilon > 0, 0 < \alpha \leq 1$,

$$Prob_{z \in Z^m}\left\{\frac{\mathcal{E}_{\mathcal{H}}(f) - \mathcal{E}_{\mathcal{H},z}(f)}{\mathcal{E}_{\mathcal{H}}(f) + \varepsilon} \geq \alpha\right\} \leq e^{-\frac{\alpha^2 m\varepsilon}{8(M^2 + M_P^2)}} \tag{75}$$

**Proof**: If $\mu = \mathcal{E}_{\mathcal{H}}(f)$, the unilateral Bernstein's inequality is applied to $\ell(f)$, and $|\ell(f)(z)| \leq M^2 + M_P^2$ is true almost everywhere. We have

$$Prob_{z \in Z^m}\left\{\frac{\mathcal{E}_{\mathcal{H}}(f) - \mathcal{E}_{\mathcal{H},z}(f)}{\mathcal{E}_{\mathcal{H}}(f) + \varepsilon} \geq \alpha\right\} \leq e^{-\frac{\alpha^2 m(\mu+\varepsilon)^2}{2\left(\sigma^2 + \frac{1}{3}(M^2 + M_P^2)\alpha(\mu+\varepsilon)\right)}} \tag{76}$$

We just have to prove the inequality.

$$\frac{\varepsilon}{8(M^2 + M_P^2)} \leq \frac{(\mu + \varepsilon)^2}{2\left(\sigma^2 + \frac{1}{3}(M^2 + M_P^2)\alpha(\mu+\varepsilon)\right)} \tag{77}$$

$$\Leftrightarrow \frac{\varepsilon}{4(M^2 + M_P^2)}\left(\sigma^2 + \frac{1}{3}(M^2 + M_P^2)\alpha(\mu + \varepsilon)\right) \leq (\mu + \varepsilon)^2 \tag{78}$$

$$\Leftrightarrow \frac{\varepsilon\sigma^2}{4(M^2 + M_P^2)} + \frac{\varepsilon\alpha\mu}{12} + \frac{\varepsilon^2\alpha}{12} \leq (\mu + \varepsilon)^2 \tag{79}$$

Since $0 < \alpha \leq 1$, bounds of the second term and the third term on the left side of the above inequality are $\varepsilon\mu$ and $\varepsilon^2$ respectively. From Lemma 9, $4max(M^2, M_P^2)\mu$ is the bound of $\sigma^2$. So, the first term is less than $\varepsilon\mu$, and $2\varepsilon\mu + \varepsilon^2 \leq (\mu + \varepsilon)^2$. □

**Lemma 11**. Given $0 < \alpha < 1$, $\varepsilon > 0$, $f \in \mathcal{H}$, we have

$$\frac{\mathcal{E}_{\mathcal{H}}(f) - \mathcal{E}_{\mathcal{H},z}(f)}{\mathcal{E}_{\mathcal{H}}(f) + \varepsilon} < \alpha \tag{80}$$

For all $g \in \mathcal{H}$, $\|f - g\|_\infty \leq \frac{\alpha\varepsilon}{4M}$, we have

$$\frac{\mathcal{E}_{\mathcal{H}}(g) - \mathcal{E}_{\mathcal{H},z}(g)}{\mathcal{E}_{\mathcal{H}}(g) + \varepsilon} < \left(\frac{3M + 2M_P}{M}\right)\alpha \tag{81}$$

**Proof**: Firstly, we have

$$\frac{\mathcal{E}_\mathcal{H}(g) - \mathcal{E}_{\mathcal{H},\mathbf{z}}(g)}{\mathcal{E}_\mathcal{H}(g) + \varepsilon} = \frac{\mathcal{E}(g) - \mathcal{E}(f_\mathcal{H}^P) + \mathcal{E}^P(g) - \mathcal{E}^P(f_\mathcal{H}^P) - \mathcal{E}_\mathbf{z}(g) + \mathcal{E}_\mathbf{z}(f_\mathcal{H}^P) - \mathcal{E}_\mathbf{z}^P(g) + \mathcal{E}_\mathbf{z}^P(f_\mathcal{H}^P)}{\mathcal{E}_\mathcal{H}(g) + \varepsilon}$$

$$= \frac{L_\mathbf{z}(g) - L_\mathbf{z}(f_\mathcal{H}^P) + L_\mathbf{z}^P(g) - L_\mathbf{z}^P(f_\mathcal{H}^P)}{\mathcal{E}_\mathcal{H}(g) + \varepsilon}$$

$$= \frac{L_\mathbf{z}(g) - L_\mathbf{z}(f)}{\mathcal{E}_\mathcal{H}(g) + \varepsilon} + \frac{L_\mathbf{z}(f) - L_\mathbf{z}(f_\mathcal{H}^P)}{\mathcal{E}_\mathcal{H}(g) + \varepsilon} + \frac{L_\mathbf{z}^P(g) - L_\mathbf{z}^P(f)}{\mathcal{E}_\mathcal{H}(g) + \varepsilon} + \frac{L_\mathbf{z}^P(f) - L_\mathbf{z}^P(f_\mathcal{H}^P)}{\mathcal{E}_\mathcal{H}(g) + \varepsilon}$$

(82)

The following inequality can be obtained from proposition 3 in literature [1] and Lemma 6.

$$\frac{\mathcal{E}_\mathcal{H}(g) - \mathcal{E}_{\mathcal{H},\mathbf{z}}(g)}{\mathcal{E}_\mathcal{H}(g) + \varepsilon} = \frac{L_\mathbf{z}(g) - L_\mathbf{z}(f)}{\mathcal{E}_\mathcal{H}(g) + \varepsilon} + \frac{L_\mathbf{z}(f) - L_\mathbf{z}(f_\mathcal{H}^P)}{\mathcal{E}_\mathcal{H}(g) + \varepsilon} + \frac{L_\mathbf{z}^P(g) - L_\mathbf{z}^P(f)}{\mathcal{E}_\mathcal{H}(g) + \varepsilon} + \frac{L_\mathbf{z}^P(f) - L_\mathbf{z}^P(f_\mathcal{H}^P)}{\mathcal{E}_\mathcal{H}(g) + \varepsilon}$$

$$\leq \frac{4M \| g - f \|_\infty}{\mathcal{E}_\mathcal{H}(g) + \varepsilon} + \frac{L_\mathbf{z}(f) - L_\mathbf{z}(f_\mathcal{H}^P)}{\mathcal{E}_\mathcal{H}(g) + \varepsilon} + \frac{8M_P \| g - f \|_\infty}{\mathcal{E}_\mathcal{H}(g) + \varepsilon} + \frac{L_\mathbf{z}^P(f) - L_\mathbf{z}^P(f_\mathcal{H}^P)}{\mathcal{E}_\mathcal{H}(g) + \varepsilon}$$

(83)

Using the first part of the proof process of Lemma 6 and $\alpha < 1$, we have

$$\left| \mathcal{E}^P(f) - \mathcal{E}^P(g) \right| \leq 4 \| f - g \|_\infty M_P \leq 4 \frac{\alpha \varepsilon}{4M} M_P = \frac{\alpha \varepsilon M_P}{M} < \frac{\varepsilon M_P}{M} \quad (84)$$

This implies

$$\mathcal{E}_\mathcal{H}^P(f) - \mathcal{E}_\mathcal{H}^P(g) = \mathcal{E}^P(f) - \mathcal{E}^P(g) \leq \frac{\varepsilon M_P}{M} \leq \mathcal{E}_\mathcal{H}^P(g) + \frac{\varepsilon M_P}{M} \quad (85)$$

And because $\mathcal{E}(f) - \mathcal{E}(f_\mathcal{H}^P) - \mathcal{E}(g) + \mathcal{E}(f_\mathcal{H}^P) \leq \varepsilon \leq \mathcal{E}(g) - \mathcal{E}(f_\mathcal{H}^P) + \varepsilon$,

If the both sides of this inequality of the both inequalities add, we have

$$\mathcal{E}(f) - \mathcal{E}(f_\mathcal{H}^P) - \mathcal{E}(g) + \mathcal{E}(f_\mathcal{H}^P) + \mathcal{E}_\mathcal{H}^P(f) - \mathcal{E}_\mathcal{H}^P(g) \leq \mathcal{E}(g) - \mathcal{E}(f_\mathcal{H}^P) + \varepsilon + \mathcal{E}_\mathcal{H}^P(g) + \frac{\varepsilon M_P}{M}$$

(86)

It is $\mathcal{E}_\mathcal{H}(f) - \mathcal{E}_\mathcal{H}(g) \leq \mathcal{E}_\mathcal{H}(g) + \frac{\varepsilon(M + M_P)}{M}$.

This is equivalent to $\dfrac{\mathcal{E}_\mathcal{H}(f) + \frac{\varepsilon(M + M_P)}{M}}{\mathcal{E}_\mathcal{H}(g) + \frac{\varepsilon(M + M_P)}{M}} \leq 2$.

Therfore,

$$\frac{L_z(f)-L_z(f_{\mathcal{H}}^P)}{\mathcal{E}_{\mathcal{H}}(g)+\varepsilon}+\frac{L_z^P(f)-L_z^P(f_{\mathcal{H}}^P)}{\mathcal{E}_{\mathcal{H}}(g)+\varepsilon}=\frac{\mathcal{E}_{\mathcal{H}}(f)-\mathcal{E}_{\mathcal{H},z}(f)}{\mathcal{E}_{\mathcal{H}}(g)+\varepsilon}<\alpha\frac{\mathcal{E}_{\mathcal{H}}(f)+\varepsilon}{\mathcal{E}_{\mathcal{H}}(g)+\varepsilon}\leq 2\alpha \quad (87)$$

So,

$$\frac{\mathcal{E}_{\mathcal{H}}(g)-\mathcal{E}_{\mathcal{H},z}(g)}{\mathcal{E}_{\mathcal{H}}(g)+\varepsilon}\leq\alpha+2\frac{M_P\alpha}{M}+2\alpha\leq\left(\frac{3M+2M_P}{M}\right)\alpha$$

$\square$

From Lemma 11, we can obtain Lemma 12.

**Lemma 12.** For all $\varepsilon>0$ and $0<\alpha<1$,

$$\text{prob}_{z\in Z^m}\left\{\sup_{f\in\mathcal{H}}\frac{\mathcal{E}_{\mathcal{H}}(f)-\mathcal{E}_{\mathcal{H},z}(f)}{\mathcal{E}_{\mathcal{H}}(f)+\varepsilon}\geq\left(\frac{3M+2M_P}{M}\right)\alpha\right\}\leq\mathcal{N}\left(\mathcal{H},\frac{\alpha\varepsilon}{4M}\right)e^{-\frac{\alpha^2 m\varepsilon}{8(M^2+M_P^2)}} \quad (88)$$

**Proof**: If $\ell=\mathcal{N}\left(\mathcal{H},\frac{\alpha\varepsilon}{4M}\right)$, there some disks $D_j, j=1,2,\dots,\ell_P$, to cover $\mathcal{H}$, which makes $f_j$ as the center and $\frac{\alpha\varepsilon}{4M}$ as its' radius. In $U$ which is a completely measurable set, we have $|f(x)-y|\leq M$. From Lemmas 11 and 6, it can be see that for all $z\in U^m$ and $f\in D_j$, we have

$$\sup_{f\in D_j}\frac{\mathcal{E}_{\mathcal{H}}(f)-\mathcal{E}_{\mathcal{H},z}(f)}{\mathcal{E}_{\mathcal{H}}(f)+\varepsilon}\geq\left(\frac{3M+2M_P}{M}\right)\alpha\Rightarrow\frac{\mathcal{E}_{\mathcal{H}}(f_j)-\mathcal{E}_{\mathcal{H},z}(f_j)}{\mathcal{E}_{\mathcal{H}}(f_j)+\varepsilon}\geq\alpha \quad (89)$$

Therefore, for $j=1,\cdots,\ell$,

$$\text{prob}_{z\in Z^m}\left\{\sup_{f\in D_j}\frac{\mathcal{E}_{\mathcal{H}}(f)-\mathcal{E}_{\mathcal{H},z}(f)}{\mathcal{E}_{\mathcal{H}}(f)+\varepsilon}\geq\left(\frac{3M+2M_P}{M}\right)\alpha\right\}\leq\text{prob}_{z\in Z^m}\left\{\frac{\mathcal{E}_{\mathcal{H}}(f_j)-\mathcal{E}_{\mathcal{H},z}(f_j)}{\mathcal{E}_{\mathcal{H}}(f_j)+\varepsilon}\geq\alpha\right\}$$
(90)

Likewise, if $\mathcal{H}=D_1\cup\cdots\cup D_\ell$ and $\varepsilon>0$, we have

$$\text{prob}_{z\in Z^m}\left\{\sup_{f\in\mathcal{H}}\frac{\mathcal{E}_{\mathcal{H}}(f)-\mathcal{E}_{\mathcal{H},z}(f)}{\mathcal{E}_{\mathcal{H}}(f)+\varepsilon}\geq\left(\frac{3M+2M_P}{M}\right)\alpha\right\}\leq\sum_{j=1}^{\ell}\text{prob}_{z\in Z^m}\left\{\sup_{f\in D_j}\frac{\mathcal{E}_{\mathcal{H}}(f)-\mathcal{E}_{\mathcal{H},z}(f)}{\mathcal{E}_{\mathcal{H}}(f)+\varepsilon}\geq\left(\frac{3M+2M_P}{M}\right)\alpha\right\} \quad (91)$$

From Lemma 10, it can be seen

$$\text{prob}_{z\in Z^m}\left\{\sup_{f\in\mathcal{H}}\frac{\mathcal{E}_{\mathcal{H}}(f)-\mathcal{E}_{\mathcal{H},z}(f)}{\mathcal{E}_{\mathcal{H}}(f)+\varepsilon}\geq\left(\frac{3M+2M_P}{M}\right)\alpha\right\}\leq\mathcal{N}\left(\mathcal{H},\frac{\alpha\varepsilon}{4M}\right)e^{-\frac{\alpha^2 m\varepsilon}{8(M^2+M_P^2)}} \quad \square$$

From Lemma 12 Theorem 5* can be proved.

In Lemma 12, $\alpha = \frac{1}{2}\left(\frac{M}{3M+2M_P}\right)$, it can be seen that in accordance with the probability of at least

$$1 - \mathcal{N}\left(\mathcal{H}, \frac{\alpha\varepsilon}{4M}\right)e^{-\frac{\alpha^2 m\varepsilon}{8(M^2+M_P^2)}} = 1 - \mathcal{N}\left(\mathcal{H}, \frac{\varepsilon}{4M}\frac{1}{2}\left(\frac{M}{3M+2M_P}\right)\right)e^{-\frac{m\varepsilon}{8(M^2+M_P^2)}\frac{1}{4}\left(\frac{M}{3M+2M_P}\right)^2} = 1 - \mathcal{N}\left(\mathcal{H}, \frac{\varepsilon}{8(3M+2M_P)}\right)e^{-\frac{m\varepsilon}{32(M^2+M_P^2)}\left(\frac{M}{3M+2M_P}\right)^2},$$
(92)

we have

$$\sup_{f\in\mathcal{H}} \frac{\mathcal{E}_{\mathcal{H}}(f)-\mathcal{E}_{\mathcal{H},z}(f)}{\mathcal{E}_{\mathcal{H}}(f)+\varepsilon} < \frac{1}{2} \tag{93}$$

So, for all $f \in \mathcal{H}$, we have $\frac{1}{2}\mathcal{E}_{\mathcal{H}}(f) < \mathcal{E}_{\mathcal{H},z}(f) + \frac{1}{2}\varepsilon$. If $f = f_z$, both sides of the above inequality are multiplied by 2, and we get

$$\mathcal{E}_{\mathcal{H}}(f_z) < 2\mathcal{E}_{\mathcal{H},z}(f_z) + \varepsilon \tag{94}$$

In according the definition of $f_z$, we have $\mathcal{E}_{\mathcal{H},z}(f_z) \leq 0$. Therefore, $\mathcal{E}_{\mathcal{H}}(f_z) < \varepsilon$. The proof of Theorem 5* is completed. □

**Proof of Theorem 6.** Firstly part (1) is proved. Since

$$\varphi(b) = \|b-a\|^2 + \tau\|b-p-\int(b-p)d\rho\|^2 + \gamma\|A^{-s}b\|^2, \tag{95}$$

if a function $\hat{b}$ minimums $\varphi$, it must be a zero point on a derivative $D\varphi$.

Suppose two operators $\Gamma(b-p) = \int(b-p)d\rho$ and $\mathcal{L} = d - \Gamma$, then we have

$$\hat{b} = [Id + \tau(Id-\Gamma)^2 + \gamma A^{-2s}]^{-1}[a + \tau(Id-\Gamma)^2 p] = (Id + \tau\mathcal{L}^2 + \gamma A^{-2s})^{-1}(a + \tau\mathcal{L}^2 p) \tag{96}$$

Because the operator $Id + \tau(Id-\Gamma)^2 + \gamma A^{-2s}$ is a sum of an identity operator and a positive definite operator, it is invertible. If $\lambda_1 \geq \lambda_2 \geq \cdots > 0$ are the eigenvalues of $A$, and function $a = \sum_{k=1}^{\infty} a_k = \sum_{k=1}^{\infty} r_k \varphi_k$, where $a_k = r_k\varphi_k$,

$$\varphi(\hat{b}) = \|\hat{b}-a\|^2 + \tau\|(Id-\Gamma)(\hat{b}-p)\|^2 + \gamma\|A^{-s}\hat{b}\|^2 \tag{97}$$

From Eq.(96), we have

$$\|\hat{b} - a\|^2 = \|(Id + \tau\mathcal{L}^2 + \gamma A^{-2s})^{-1}[\tau\mathcal{L}^2 p - (\tau\mathcal{L}^2 + \gamma A^{-2s})a]\|^2 \tag{98}$$

And we get

$$\|(Id - \Gamma)(\hat{b} - p)\|^2 = \|\mathcal{L}(Id + \tau\mathcal{L}^2 + \gamma A^{-2s})^{-1}(a - p - \gamma A^{-2s}p)\|^2 \tag{99}$$

and

$$\begin{aligned}
\|A^{-s}\hat{b}\|^2 &= \|A^{-s}(Id + \tau\mathcal{L}^2 + \gamma A^{-2s})^{-1}(a + \tau\mathcal{L}^2 p)\|^2 \\
&= \|(Id + \tau\mathcal{L}^2 + \gamma A^{-2s})^{-1}(A^{-s}a + \tau A^{-s}\mathcal{L}^2 p)\|^2 \\
&= \sum_{k=1}^{\infty}(1 + \tau\mathcal{L}^2 + \gamma\lambda_k^{-2s})^{-2}(\lambda_k^{-s}a_k + \tau\lambda_k^{-s}\mathcal{L}^2 p_k)^2 \\
&= \sum_{k=1}^{\infty}\lambda_k^{2r-2s}(1 + \tau\mathcal{L}^2 + \gamma\lambda_k^{-2s})^{-2}\lambda_k^{-2r}(a_k + \tau\mathcal{L}^2 p_k)^2 \\
&\leq \sup_{t\in\mathbb{R}} t^{2r-2s}(1 + \tau\mathcal{L}^2 + \gamma t^{-2s})^{-2} \sum_{k=1}^{\infty}\lambda_k^{-2r}(a_k + \tau\mathcal{L}^2 p_k)^2
\end{aligned}$$

Then we have

$$(2r - 2s)\hat{t}^{2r-2s-1}(1 + \tau\mathcal{L}^2 + \gamma\hat{t}^{-2s})^{-2} + 4s\gamma\hat{t}^{2r-4s-1}(1 + \tau\mathcal{L}^2 + \gamma\hat{t}^{-2s})^{-3} = 0 \tag{100}$$

Finally, $\hat{t} = (r+s)^{\frac{1}{2s}}\gamma^{\frac{1}{2s}}(s-r)^{-\frac{1}{2s}}(1+\tau\mathcal{L}^2)^{-\frac{1}{2s}}$.

Then $\psi(\hat{t}) = \hat{t}^{2r-2s}(1 + \tau\mathcal{L}^2 + \gamma\hat{t}^{-2s})^{-2} \leq (r+s)^{\frac{r+s}{s}}\gamma^{\frac{r-s}{s}}(s-r)^{-\frac{r+s}{s}}(1+\tau\mathcal{L}^2)^{-\frac{r+s}{s}}$.

$$\|A^{-s}\hat{b}\|^2 \leq \psi(\hat{t})\|A^{-r}(a + \tau\mathcal{L}^2 p)\|^2. \tag{101}$$

$\varphi(\hat{b}) \leq$
$\|(Id + \tau\mathcal{L}^2 + \gamma A^{-2s})^{-1}[\tau\mathcal{L}^2 p - (\tau\mathcal{L}^2 + \gamma A^{-2s})a]\|^2 + \tau\|\mathcal{L}(Id + \tau\mathcal{L}^2 + \gamma A^{-2s})^{-1}(a - p - \gamma A^{-2s}p)\|^2 + (r+s)^{\frac{r+s}{s}}\gamma^{\frac{r-s}{s}}(s-r)^{-\frac{r+s}{s}}(1+\tau\mathcal{L}^2)^{-\frac{r+s}{s}}\|A^{-r}(a + \tau\mathcal{L}^2 p)\|^2. \tag{102}$

For part (2), if $\|A^{-s}a\| \leq R$, $\|A^{-s}p\| \leq R$ and $a = p$, the minimum value of the expression is zero. Then the theorem is obviously true. Suppose the case is not true, in the subspace $\|A^{-s}b\| \leq R$ of $\mathcal{H}$, the optimal $\hat{b}$ is on the boundary of the subspace, which is $\|A^{-s}\hat{b}\| = R$. From Eq.(101), we have

$$\gamma^{\frac{s-r}{s}} \leq (r+s)^{\frac{r+s}{s}} R^{-2}(s-r)^{-\frac{r+s}{s}}(1+\tau \mathcal{L}^2)^{-\frac{r+s}{s}} \|A^{-r}(a+\tau \mathcal{L}^2 p)\|^2 \tag{103}$$

Then, we get

$$\gamma \leq (r+s)^{\frac{r+s}{s-r}} R^{-\frac{2s}{s-r}}(s-r)^{-\frac{r+s}{s-r}}(1+\tau \mathcal{L}^2)^{-\frac{r+s}{s-r}} \|A^{-r}(a+\tau \mathcal{L}^2 p)\|^{\frac{2s}{s-r}} \qquad \square$$

## 9. Experimental evaluation

Experiments were performed to evaluate the generalization performance and interpretability of the proposed ILSSVM algorithm for proving that the generalization performance is not reduced by the improved interpretability and that the interpretation model can be close enough to original model according to the interpretation distance. The Gaussian kernel (RBF) was used in the experiments. Nelder–Mead simplex optimization algorithm[23], Multi-objective genetic algorithm and 10-fold cross-validation were employed for hyper-parameter optimization. All methods were implemented in MATLAB R2009a and run on a PC with an Intel i73520M 2.90-GHz CPU and 4 GB of RAM. The mean square error (MSE) and the Pearson product-moment correlation coefficient (PPCC) were used to estimate the generalization ability and the interpretability distance (ID) and PPCC were used to estimate the interpretability. The lower the MSE value is, the closer the predictive value is to the target value. In the same way, the lower the ID value is, the power the interpretability of the predictive model is. The closer the value of PPCC is to 1, the better the predictive value and the target value correlate with each other. In the experiments, all data are normalized within $[-1,1]$ to evaluate the generalization performance and interpretability.

### 9.1 Data sets

Benchmark data sets for the experiments were generated using four benchmark functions: the Friedman, Plane, Multi, and Gabor functions. These benchmark data sets were assigned various forms of interference for the overall testing of the generalization performance and interpretability of the both algorithms, such as various forms of noise and nuisance attributes.

The Friedman benchmark data set is a common nonlinear regression data set with 50 samples and was generated by the Friedman function $y = 10sin(\pi x_1 x_2) + 20(x_3 - 0.5)^2 + 10x_4 + 5x_5 + \varepsilon$. The Gaussian random number $\varepsilon$ was assigned to 30 random samples as a random number from the normal distribution $N(0,1)$ and to another 30 samples as a random number from the normal distribution $N(0,5)$. The data set is denoted by "Friedman1". Meanwhile, the first 20 samples of the data set were generated with an additional Gaussian noise of mean 0 and standard deviation 1, as were the second 20 samples with an additional Gaussian noise of mean 0 and standard deviation 2 and were the final 20 samples with an additional Gaussian noise of mean 0 and standard deviation 3. The data set is denoted by "Friedman2".

The Plane benchmark data set is a common linear regression data set with 60 samples and was generated by the plane function $y = 0.6x_1 + 0.3x_2, -1 < x_1, x_2 < 1$. All samples of the data set were generated with an additional Gaussian noise of mean 0 and standard deviation 1. The data set is denoted by "Plane1". Then all samples were added one random number of mean 0 and standard deviation 1 as a new attribution. The new data set is denoted by "Plane2".

The Multi benchmark data set is a common nonlinear regression data set and was generated by the multi-function $y = 0.79 + 1.27x_1x_2 + 1.56x_1x_4 + 3.42x_2x_5 + 2.06x_3x_4x_5$, $-1 < x_1, x_2, x_3, x_4, x_5 < 1$. The data set includes 60 samples. All samples had a Gaussian noise of mean 0 and standard deviation 1 added. The data set is denoted by "Multi1". Then all samples were added 5 random numbers of mean 0 and standard deviation 1 as new 5 attributions. The new data set is denoted by "Multi2".

The Gabor benchmark data set is a common nonlinear regression data set with 60 samples generated by the Gabor function $= \pi e^{-2(x_1^2 + x_2^2)} cos(2\pi(x_1 + x_2))/2$, $-1 < x_1, x_2 < 1$. For the overall testing, all samples had a Gaussian noise of mean 0 and standard deviation 1 added. The data set is denoted by "Gabor1". Meanwhile, the first 30 samples had a Gaussian noise of mean 0 and standard deviation 0.1 added, and the other 30 samples had a Gaussian noise of mean 0 and standard deviation 0.5 added. The data set is denoted by "Gabor2".

## 9.2 Performance test and discussion

### 9.2.1 Generation of the interpretability model

All optimal interpretability models are found by Particle Swarm Optimization algorithm (PSO) in which interpretability distance between interpretability model and generation function of these benchmark data sets is used as its fitness function. All interpretability models are constructed by the method in our paper [14].

For every data set, 10 experiments were used to evaluate or compare its interpretability model and its generation function, and the interpretability distances of each experiment and their averages were computed, as shown under test estimations in Table 1.

Table 1 the interpretability distances between its interpretability model and its generation function

| Dataset | 1 | 2 | 3 | 4 | 5 | 6 | 7 | 8 | 9 | 10 | Average |
|---|---|---|---|---|---|---|---|---|---|---|---|
| Mutil1 | 0.0601 | 0.09106 | 0.089386 | 0.088967 | 0.116376 | 0.117979 | 0.073959 | 0.13071 | 0.104156 | 0.08413 | 0.095682 |
| Mutil2 | 0.077815 | 0.191737 | 0.107417 | 0.144015 | 0.069214 | 0.055446 | 0.112679 | 0.11743 | 0.116744 | 0.119284 | 0.111178 |
| Friedman1 | 0.15129 | 0.089082 | 0.130954 | 0.02929 | 0.042874 | 0.164895 | 0.174471 | 0.143432 | 0.182622 | 0.192851 | 0.130176 |
| Friedman2 | 0.21387 | 0.206628 | 0.179054 | 0.033768 | 0.174034 | 0.136288 | 0.152859 | 0.030009 | 0.115551 | 0.198206 | 0.144027 |
| Gabor1 | 0.048825 | 0.047 | 0.16672 | 0.038508 | 0.079201 | 0.234266 | 0.05823 | 0.094606 | 0.047894 | 0.042043 | 0.085729 |
| Gabor2 | 0.061795 | 0.03271 | 0.040506 | 0.049996 | 0.038837 | 0.086372 | 0.047522 | 0.049858 | 0.156116 | 0.156997 | 0.072071 |
| Plane1 | 0.116733 | 0.139379 | 0.136406 | 0.119136 | 0.142982 | 0.01674 | 0.157598 | 0.137866 | 0.156004 | 0.098645 | 0.122149 |
| Plane2 | 0.469269 | 0.111113 | 0.200923 | 0.202838 | 0.140311 | 0.112811 | 0.161657 | 0.537809 | 0.186284 | 0.617176 | 0.275019 |

As seen from Table 1, all interpretability distances and their average values are enough low. These results prove that these interpretability models have enough correct interpretation about these data sets and the method of constructed interpretability model is feasible.

### 9.2.2 Performance test

The two algorithms, i.e., ILSSVM, and Least-squares Support Vector Machine (LSSVM), were applied to learning on the nine benchmark data sets, namely the Friedman1, Friedman2,Plane1, Plane2, Multi1, Multi2,Gabor1, Gabor1, and toy data sets, based on RBF kernels. For every data set, 10-fold cross-validation was used to evaluate or compare the both learning algorithms. The average values (namely, R_ID, MSE and SCC), variances and standard deviations of the interpretability distance, MSE and PPCC of each experiment were computed, as shown in Table 2. In every cross-validation, the interpretability of these models also were evaluated by computing the MSE and SCC between the prediction result and the noise-free result from the generation functions of these data sets. Their average results (namely, R_MSE and R_SCC), variances and standard deviations are shown in Table 2. The time estimations are evaluations of the time costs of the two algorithms.

Table 2 testing results on the benchmark data sets (the best values are underlined, and the second-best values are given in italics.)

| Dataset | Method | MSE | SCC | R_ID | R_MSE | R_SCC | D_time |
|---|---|---|---|---|---|---|---|
| Mutil1 | LSSVM | 0.114427 | **0.5433857** | 0.1065015 | 0.1330839 | **0.2333565** | **0.019125** |
| | VAR | 0.003697 | 0.066281 | 0.009031 | 0.010135 | 0.062756 | 0.000003 |
| | STD | 0.060807 | 0.257452 | 0.09503 | 0.100671 | 0.250511 | 0.001763 |
| | ILSSVM | **0.1076508** | 0.4817993 | **0.075098** | **0.0916694** | 0.1951125 | 0.0253975 |
| | VAR | 0.002539 | 0.069099 | 0.004818 | 0.002709 | 0.048812 | 0.00002 |
| | STD | 0.050393 | 0.262867 | 0.069413 | 0.052044 | 0.220935 | 0.004447 |
| Mutil2 | LSSVM | **0.2172519** | 0.4086633 | 0.1112864 | 0.1068857 | **0.2650952** | **0.0215691** |
| | VAR | 0.024391 | 0.075867 | 0.006345 | 0.004045 | 0.075213 | 0.000003 |
| | STD | 0.156177 | 0.275439 | 0.079655 | 0.063604 | 0.27425 | 0.001689 |
| | ILSSVM | 0.262797 | **0.434627** | **0.014717** | **0.022123** | 0.204703 | 0.027529 |
| | VAR | 0.019235 | 0.059991 | 0.000172 | 0.000158 | 0.045133 | 0.000012 |
| | STD | 0.13869 | 0.244931 | 0.013101 | 0.012555 | 0.212445 | 0.003423 |
| Friedman1 | LSSVM | 0.1067997 | 0.4253001 | 0.0430994 | 0.0655943 | 0.1649695 | **0.0193826** |
|  | VAR | 0.009574 | 0.099418 | 0.000493 | 0.001766 | 0.036338 | 0.000014 |
|  | STD | 0.097846 | 0.315307 | 0.022205 | 0.042025 | 0.190625 | 0.003781 |
|  | ILSSVM | **0.1049383** | **0.4442116** | **0.0197024** | **0.0396459** | **0.194748** | 0.0237394 |
|  | VAR | 0.009527 | 0.071058 | 0.000199 | 0.001089 | 0.081116 | 0.000004 |
|  | STD | 0.097608 | 0.266567 | 0.014109 | 0.032994 | 0.284808 | 0.00189 |
| Friedman2 | LSSVM | **0.0500114** | **0.6926092** | 0.0949282 | 0.1285254 | **0.3679689** | **0.0197053** |
|  | VAR | 0.002169 | 0.086844 | 0.003976 | 0.004089 | 0.098477 | 0.000005 |
|  | STD | 0.046569 | 0.294693 | 0.063056 | 0.063942 | 0.313811 | 0.002254 |
|  | ILSSVM | 0.0708551 | 0.6856288 | **0.049728** | **0.0822261** | 0.3288525 | 0.024227 |
|  | VAR | 0.002606 | 0.104654 | 0.000782 | 0.002228 | 0.077344 | 0.000005 |
|  | STD | 0.051047 | 0.323502 | 0.027967 | 0.047197 | 0.278109 | 0.002304 |
| Gabor1 | LSSVM | 0.1378119 | **0.5917432** | 0.0523839 | 0.0579846 | **0.2614655** | **0.0154235** |
|  | VAR | 0.075219 | 0.087038 | 0.006758 | 0.008447 | 0.080128 | 0.000002 |
|  | STD | 0.274261 | 0.295022 | 0.08221 | 0.091906 | 0.283068 | 0.001411 |
|  | ILSSVM | **0.0921511** | 0.5726303 | **0.0301368** | **0.0319931** | 0.2202061 | 0.0224847 |
|  | VAR | 0.015144 | 0.119472 | 0.001249 | 0.001233 | 0.07961 | 0.000036 |
|  | STD | 0.123061 | 0.345647 | 0.035341 | 0.03511 | 0.282152 | 0.005985 |
| Gabor2 | LSSVM | **0.0958341** | 0.6599208 | 0.1243709 | 0.2091739 | **0.3556074** | 0.0145596 |
|  | VAR | 0.00469 | 0.083998 | 0.00594 | 0.005074 | 0.090976 | 0 |
|  | STD | 0.068484 | 0.289824 | 0.077072 | 0.071229 | 0.301623 | 0.000564 |
|  | ILSSVM | 0.2255035 | 0.2864574 | **0.0775727** | **0.1533118** | 0.2682881 | 0.0207213 |
|  | VAR | 0.024788 | 0.1518673 | 0.012052 | 0.006113 | 0.0733818 | 0.00001 |
|  | STD | 0.157442 | 0.3674142 | 0.109781 | 0.078187 | 0.2553983 | 0.003158 |
| Plane1 | LSSVM | **0.0182005** | **0.9160869** | 0.1188416 | 0.1301738 | **0.9902211** | **0.0149569** |
|  | VAR | 0.000169 | 0.005333 | 0.005154 | 0.00668 | 0.000063 | 0.000001 |
|  | STD | 0.012989 | 0.073028 | 0.07179 | 0.081732 | 0.007918 | 0.000778 |
|  | ILSSVM | 0.0527447 | 0.8934982 | **0.058342** | **0.0753191** | 0.9477858 | 0.0194409 |
|  | VAR | 0.00339 | 0.009372 | 0.001533 | 0.002684 | 0.006082 | 0.003402 |
|  | STD | 0.058223 | 0.096808 | 0.039153 | 0.051809 | 0.077987 | 0.058325 |
| Plane2 | LSSVM | **0.0324038** | **0.7672276** | 0.1195876 | 0.1391647 | 0.4549694 | **0.0154218** |
|  | VAR | 0.000526 | 0.100921 | 0.012331 | 0.010013 | 0.084897 | 0 |
|  | STD | 0.022933 | 0.317681 | 0.111046 | 0.100066 | 0.291372 | 0.000656 |
|  | ILSSVM | 0.0992704 | 0.6110135 | **0.0559536** | **0.0863349** | **0.4630502** | 0.0197023 |
|  | VAR | 0.007281 | 0.112323 | 0.006174 | 0.00563 | 0.123061 | 0.000001 |
|  | STD | 0.085328 | 0.335146 | 0.078575 | 0.075032 | 0.350801 | 0.001044 |

As seen from Table 2, for the Mutil1, Friedman1 and Gabor1, the average MSE values of ILSSVM are often significantly lower than those of LSSVM, and the average SCC values of ILSSVM are not obviously improved compared with those of LSSVM. But the average interpretability distances (R_ID) of ILSSVM and the MSE values (R_MSE) between the prediction result and the noise-free result from the generation functions of these data sets of ILLSVM are obviously the most improved compared to those of LSSVM. At the same time, the SCC values (R_SCC) between the prediction result and the noise-free result from the generation functions of these data sets of ILLSVM are significant drop compared to those of LSSVM. These results might also prove that ILSSVM can converge to the global optimum more effectively than LSSVM for excellent generalization performance on all homoscedasticity data sets, while the constraints in ILSSVM can tune the shape of the regression models for good interpretability.

For the Mutil2, Friedman2, Gabor2, Plane1 and Plane2, the average MSE values of LSSVM are often significantly lower than those of ILSSVM, and the average PPCC values of LSSVM are not obviously better than those of ILSSVM except for Gabor2. However, it can be clearly seen that the average interpretability distances (R_ID) of ILSSVM are obviously the most improved compared to those of LSSVM. At the same time, we can find that the MSE values (R_MSE) between the prediction result and the noise-free result from the generation functions of these data sets of ILLSVM also are significant less than those of LSSVM. The results indicate that the MSE value can also estimate the interpretability performance of the model after all the data are normalized and the effect of different magnitudes of all data is eliminated. This means indirectly that the definition of the interpretation distance is reasonable. These results may prove that ILSSVM has excellent generalization performance and excellent interpretability than LSSVM all heteroscedastic data sets or data sets with unrelated attributes.

## 10. Conclusion

In this paper, to improve the interpretability of a predictive model we proposed a quantitative index of the interpretability, and analyzed the relationship between the interpretability and the generalization performance of the prediction model in machine learning. The sufficient and necessary condition for consistent convergence of expected risk and interpretation distance was derived based on the definition of the consistent convergence. Optimal error bound of hypothesis space was discussed when the interpretability and the generalization performance of the prediction model are concerned together. The learning framework for improving the interpretability is proposed and the condition for global optimal solution based on the framework was deduced. The learning framework was applied to the least-squares support vector machine and was evaluated by some experiments. The following conclusions can be drawn.

In practical engineering, the paper provides a new learning paradigm. When we input a data set and some fuzzy relationships between attributes, the paradigm outputs a model with interpretability and prediction abilities. The model can not only predict an accurate result for new input data but also provide a reasonable causal relationship between the result and the input data which is expressed by the interpretability model. Thus, the interpretability of the prediction model is improved.

## Acknowledgments

We would like to acknowledge support for this project from the National Natural Science Foundation of China (Program Numbers 61672027, 61773314, 61773313), the National Key R&D Program of China(Program Number 2017YFB1201500), the Natural Science Foundation of Shaanxi Province (Program Numbers 2014JQ8299, 2017JM6080) and the Specialized Research Fund for the Doctoral Program of Higher Education (Program Number 20136118120011).